\documentclass{article}

\usepackage[preprint]{neurips_2026}

\usepackage[utf8]{inputenc} % allow utf-8 input
\usepackage[T1]{fontenc}    % use 8-bit T1 fonts
\usepackage{hyperref}       % hyperlinks
\usepackage{url}            % simple URL typesetting
\usepackage{booktabs}       % professional-quality tables
\usepackage{amsfonts}       % blackboard math symbols
\usepackage{nicefrac}       % compact symbols for 1/2, etc.
\usepackage{microtype}      % microtypography
\usepackage{xcolor}         % colors

\usepackage{amsmath, amssymb, amsthm}
\usepackage{graphicx}
\usepackage{bbm}          % \mathbbm{1}
\usepackage{algorithm}
\usepackage{algpseudocode}
\usepackage{multirow}
\usepackage{natbib}
\usepackage{pifont}  % for \ding{55} (cross mark)
\usepackage{wrapfig} % inline figures wrapped by text

\newcommand{\vN}[2]{$#1_{\pm#2}$}
\newcommand{\vA}[2]{$\mathbf{#1}_{\pm#2}$}
\newcommand{\vB}[2]{$\underline{#1}_{\pm#2}$}

\title{\emph{DRIFT}: A Benchmark for Task-Free Continual Graph Learning with Continuous Distribution Shifts}
\author{%
Guiquan Sun$^{1}$,~
Xikun Zhang$^{2}$\thanks{Corresponding author.},~
Jingchao Ni$^{3}$,~
Dongjin Song$^{1}$\footnotemark[1] \\
\textsuperscript{1}University of Connecticut,~\textsuperscript{2}RMIT University,~\textsuperscript{3}University of Houston
}

\begin{document}

\maketitle

\begin{abstract}
% Continual graph learning (CGL) aims to learn from dynamically evolving graphs while mitigating catastrophic forgetting. Existing CGL approaches typically adopt a task-based formulation, where the data stream is partitioned into a sequence of discrete tasks with pre-defined boundaries. However, such assumptions rarely hold in real-world environments, where data distributions evolve continuously and task identity is often unavailable. To address this gap, we reformulate continual graph learning from a task-free perspective that better reflects real-world non-stationary environments. We propose a unified formulation that models the data stream as a time-varying mixture of latent task distributions, enabling a continuous characterization of distribution drift. To simulate realistic non-stationary environments, we construct a new benchmark, \emph{DRIFT}, that spans a spectrum of transition dynamics ranging from hard task switches to smooth distributional shifts using Gaussian mixing curves. We evaluate representative continual learning methods under this task-free setting and observe significant performance degradation compared to traditional task-based evaluation protocols. Our findings indicate that many existing approaches implicitly rely on task boundary information and may not generalize well to realistic task-free streaming scenarios. This work highlights the importance of studying continual graph learning under more realistic non-stationary conditions and provides a benchmark for future research in this direction.

Continual graph learning (CGL) aims to learn from dynamically evolving graphs while mitigating catastrophic forgetting. Existing CGL approaches typically adopt a task-based formulation, where the data stream is partitioned into a sequence of discrete tasks with pre-defined boundaries. However, such assumptions rarely hold in real-world environments, where data distributions evolve continuously and task identity is often unavailable. To better reflect realistic non-stationary environments, we revisit continual graph learning from a task-free perspective. We propose a unified formulation that models the data stream as a time-varying mixture of latent task distributions, enabling continuous modeling of distribution drift. Based on this formulation, we construct \emph{DRIFT}, a benchmark that spans a spectrum of transition dynamics ranging from hard task switches to smooth distributional drift through a Gaussian parameterization. We evaluate representative continual learning methods under this task-free setting and observe substantial performance degradation compared to traditional task-based protocols. Our findings indicate that many existing approaches implicitly rely on task boundary information and struggle under realistic task-free graph streams. This work highlights the importance of studying continual graph learning under realistic non-stationary conditions and provides a benchmark for future research in this direction. Our code is available at \url{https://github.com/UConn-DSIS/DRIFT}.
\end{abstract}

\section{Introduction}
\label{sec:intro}

Continual learning on graphs (CGL) has emerged as an important research direction for modeling dynamic relational data, with applications in recommendation systems, social networks, financial modeling, and evolving knowledge graphs.
Real-world graph data is inherently non-stationary: user-item interaction networks evolve as preferences change over time~\cite{hamilton2017sage}, citation networks continuously incorporate new papers and topics~\cite{hu2020open}, and knowledge graphs expand with emerging entities and relations~\cite{wang2017knowledge}.
In these scenarios, models must continuously adapt to new information while preserving previously acquired knowledge, making continual learning a natural paradigm for graph-based systems.

Despite this motivation, most existing CGL studies adopt a task-based formulation, where the data stream is artificially partitioned into a sequence of discrete tasks.
This formulation typically assumes that task boundaries are known during training, and that transitions between tasks are abrupt, meaning the data distribution switches instantaneously from one task to another.
Such assumptions simplify evaluation and enable controlled measurement of catastrophic forgetting, and have been widely adopted in continual learning benchmarks~\cite{kirkpatrick2017overcoming, chaudhry2019rs} as well as continual graph learning benchmarks such as CGLB~\cite{zhang2022cglb}.
However, these assumptions rarely hold in realistic environments.

In real-world systems, data distributions typically evolve continuously rather than discretely.
For example, in recommendation systems, user preferences gradually shift as new interests emerge while previously relevant items remain partially informative~\cite{hamilton2017sage}.
In citation networks, research topics evolve over time with overlapping periods of influence rather than abrupt transitions~\cite{hu2020open}.
Temporal graph learning studies further show that relational structures evolve progressively as new edges appear and node representations drift~\cite{rossi2020temporal}.
Similarly, knowledge graphs and interaction networks continuously incorporate new entities and relations without explicit task boundaries~\cite{wang2017knowledge}.
In these settings, task identity is typically unobservable, and distribution shifts occur gradually over time.

These observations suggest that continual graph learning should be modeled as learning under non-stationary streaming distributions rather than as a sequence of isolated tasks.
However, existing benchmarks largely rely on hard task transitions and explicit task identity, making it unclear whether current methods can handle more realistic distribution dynamics.
While recent work has explored blurry or stochastic task transitions in image-based continual learning~\cite{koh2022online,moon2023online}, the effect of transition structure on continual graph learning remains underexplored.

In this work, we revisit continual graph learning from a task-free perspective.
We model data streams as time-varying mixtures of latent task distributions, enabling a continuous characterization of distribution drift.
To simulate realistic transitions, we adopt a Gaussian mixing curve that produces smooth and localized changes in distribution dominance over time.
This choice is motivated by the observation that many real-world transitions arise from the accumulation of multiple independent factors, resulting in gradual changes in data distributions.
The Gaussian formulation further provides a simple and controllable parameterization of temporal overlap, enabling a unified framework that spans from hard task switches to fully continuous drift.

Based on this formulation, we introduce \emph{DRIFT}, a benchmark for task-free continual graph learning that systematically varies the degree of distribution overlap.
Through extensive experiments, we evaluate representative continual learning methods and observe substantial performance degradation when task identity is unavailable or when transitions become smooth.
These results suggest that many existing approaches implicitly rely on unrealistic boundary assumptions and may not generalize well to real-world streaming environments.

Our \textbf{contributions} are summarized as follows:
(\textbf{C1}) We identify limitations of existing task-based CGL formulations and argue that continual graph learning should be studied under task-free streaming settings.
(\textbf{C2}) We introduce a unified formulation that models task transitions as time-varying mixture distributions.
(\textbf{C3}) We propose a benchmark that spans a continuous spectrum of transition dynamics via Gaussian mixing curves.
(\textbf{C4}) We empirically demonstrate that existing continual learning methods degrade significantly under realistic non-stationary settings.
\vspace{-3mm}
\section{Related Work}
\label{sec:related}
\vspace{-2mm}
\textbf{Continual Learning}
Continual learning methods are typically grouped into rehearsal-based approaches~\cite{chaudhry2019rs,chaudhry2018efficient,aljundi2019gradient,buzzega2020dark}, regularization-based methods~\cite{kirkpatrick2017overcoming,aljundi2019task,li2017learning}, and architecture-based strategies~\cite{zhang2023continual}. Most of these methods assume explicit task identities and well-defined task boundaries. To relax this assumption, recent works study task-free and online settings, including blurry task boundaries (i-Blurry~\cite{koh2022online}, Si-Blurry~\cite{moon2023online}) and fully online protocols without task segmentation~\cite{mai2022online}. However, existing benchmarks are largely based on image classification, where samples are independent. As a result, they do not capture structured dependencies or transition dynamics in relational data. In contrast, we study how transition structure governs continual learning behavior, focusing on continuous distribution shifts rather than discrete task boundaries.

\textbf{Continual Graph Learning}
Continual graph learning (CGL) extends continual learning to relational data~\cite{han2024topology,zhang2024topology,zhang2022hierarchical,sun2025hero,liu2023cat,wang2022lifelong,niu2024replay,tian2026class,li2024matters}, 
introducing challenges such as inter-task edges, evolving neighborhoods, and 
topology shift. Existing methods incorporate graph-specific mechanisms, including 
replay-based approaches (e.g., ERGNN~\cite{zhou2021overcoming} and ContinualGNN~\cite{wang2020streaming}), topology-aware 
regularization (TWP~\cite{liu2021overcoming}), structural memory compression 
(SSM~\cite{zhang2022sparsified}, SEM~\cite{zhang2023ricci}), and diversity-driven memory sampling(DMSG~\cite{qiao2025towards}). Despite these advances, most CGL benchmarks (e.g., CGLB~\cite{zhang2022cglb} and Zhang et al.~\cite{zhang2024continual}) adopt 
task-based protocols with hard transitions and explicit task identities, leading 
to abrupt distribution shifts. Recent online CGL settings~\cite{donghi2025online} 
remove multiple passes over data, but still do not explicitly model transition 
dynamics. In contrast, our work focuses on continuous distribution evolution, 
providing a unified framework to study how different transition structures 
(e.g., hard, blurry, and smooth) affect continual learning on graphs.

\textbf{Dynamic and Temporal Graph Learning}
Dynamic graph learning models temporal evolution via time-stamped edges or graph snapshots, with representative approaches such as TGN~\cite{rossi2020temporal}, TGAT~\cite{xu2020inductive}, and DySAT~\cite{sankar2020dysat} capturing evolving node representations and temporal dependencies. While these methods address non-stationarity, they do not explicitly consider the continual learning objective of retaining performance on previously observed distributions. In contrast, continual graph learning focuses on the adaptation-retention trade-off under distribution shifts. Our work connects these perspectives by introducing transition structures inspired by gradual drift and models distribution evolution as a continuous process rather than assuming discrete task boundaries.

\begin{figure}[t]
    \centering
    \includegraphics[width=0.99\columnwidth]{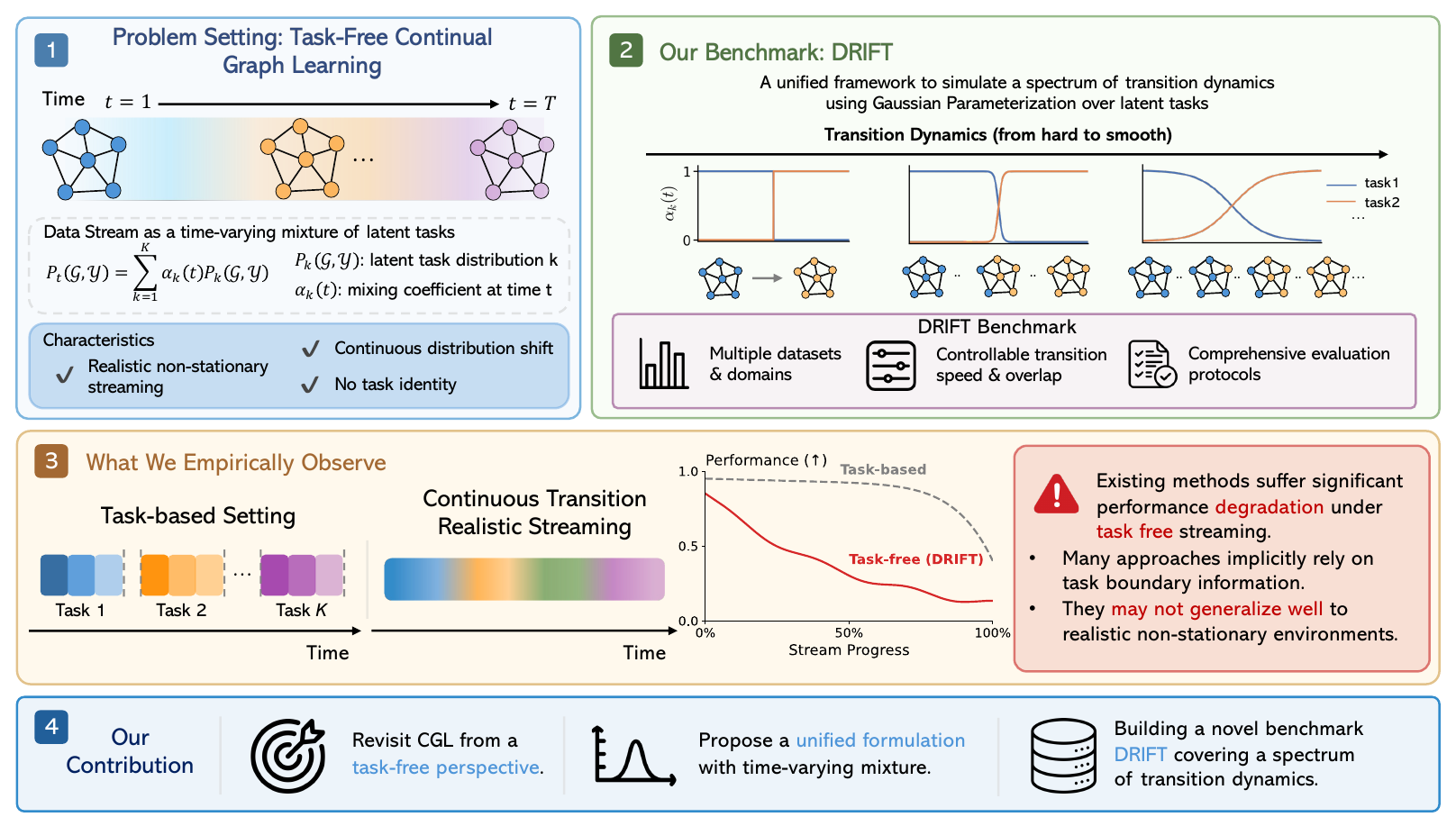}
    \vspace{-2mm}
    \caption{Overview of DRIFT. We propose a unified formulation of task-free continual graph learning based on a time-varying mixture of latent tasks, and construct a benchmark with controllable transition dynamics via Gaussian parameterization. This enables systematic evaluation across transition regimes, revealing significant performance degradation of existing methods under realistic streaming.}
    \label{fig:overview}
\vspace{-3mm}
\end{figure}

\vspace{-2mm}
\section{Problem Formulation}
\label{sec:prob_form}

\subsection{Task-based Continual Graph Learning}

In the standard continual graph learning setting, the data stream is divided into a sequence of tasks
\[
\mathcal{T}_1, \mathcal{T}_2, \dots, \mathcal{T}_K,
\]
where each task $\mathcal{T}_k$ is associated with a dataset $\mathcal{D}_k$ drawn from a task-specific distribution.

The model is trained sequentially on each task, and the objective is to learn a function $f_\theta$ that performs well on all tasks while mitigating catastrophic forgetting.

A key assumption in this formulation is the availability of task identity. At each training step, the model is aware of which task the current data belongs to, either explicitly (via task labels) or implicitly (via known task boundaries). Moreover, transitions between tasks are typically modeled as abrupt switches:
\[
\mathcal{D}(t) =
\begin{cases}
\mathcal{D}_1, & t \in [t_1, t_2) \\
\mathcal{D}_2, & t \in [t_2, t_3) \\
\cdots
\end{cases}
\]
Such a formulation implies that different tasks are non-overlapping and strictly separated in time.

\subsection{Task-Free Continual Graph Learning}

We instead consider a task-free setting~\cite{aljundi2019task} where data arrives as a continuous stream of mini-batches
\[
\{\mathcal{B}_1, \mathcal{B}_2, \dots, \mathcal{B}_T\},
\]
without any observable task identity or boundary information.

At each time step $t$, the data is sampled from a time-varying distribution $\mathcal{D}_t$, which evolves continuously over time. We model this distribution as a mixture of latent task distributions:
\begin{equation}
\mathcal{D}_t = \sum_{k=1}^{K} \alpha_k(t) \, \mathcal{D}_k,
\end{equation}
where $\alpha_k(t) \in [0,1]$ and $\sum_k \alpha_k(t) = 1$.
Here, $\alpha_k(t)$ represents the contribution of the $k$-th latent task at time $t$. Unlike the task-based setting, task identity is not observable, and the model must adapt purely based on the incoming data stream. This formulation generalizes traditional CGL: the task-based setting corresponds to the special case where $\alpha_k(t) \in \{0,1\}$.

\subsection{A Unified View of Task Transitions}

We provide a unified formulation of task transitions through a time-dependent mixing function $\alpha_k(t)$, which specifies the contribution of each latent task $k$ at time $t$.

\textbf{Hard Transition.}
The conventional setting corresponds to a piecewise constant function $\alpha_k(t) \in \{0,1\}$, where only one task is active at any time.

\textbf{Boundary-Local Mixing.}
To model blurry task boundaries, adjacent tasks are mixed within a limited transition window, resulting in local overlap near boundaries (i.e., $\alpha_k(t) > 0$ for $k = \{n, n+1\}, t \in [t_1, t_2]$).

\textbf{Global Mixing.}
All tasks remain active throughout the stream with small but non-zero probabilities (i.e., $\alpha_k(t) > 0$ for all $k,t$), inducing persistent background distributions.

\textbf{Continuous Gaussian Transition.}
We model smooth distribution drift by parameterizing $\alpha_k(t)$ with a Gaussian kernel centered at $\mu_k$:
\begin{equation}
    \alpha_k(t) = \frac{\exp\left(-\frac{(t - \mu_k)^2}{2\sigma^2}\right)}
    {\sum_j \exp\left(-\frac{(t - \mu_j)^2}{2\sigma^2}\right)},
\end{equation}
where $\sigma$ controls the degree of temporal overlap across tasks. Larger $\sigma$ yields smoother transitions, while smaller $\sigma$ recovers near-discrete task switches.

In practice, we realize this process at the mini-batch level by sampling instances according to $\alpha_k(t)$ to construct each batch.
\vspace{-1mm}
\section{Benchmark Construction}
\label{sec:benchmark}

Inspired by prior work that simulates task-free streams from static datasets~\cite{chrysakis2023simulating}, we construct a task-free continual graph learning benchmark based on the unified mixture formulation introduced in Section~\ref{sec:prob_form}. Our goal is to simulate non-stationary data streams under a unified and controllable transition framework. Rather than defining separate protocols for different settings, we instantiate this framework with a simple yet effective parameterization that enables continuous control over transition dynamics. An overview of our benchmark is shown in Figure~\ref{fig:overview}.

\subsection{Task Decomposition}

Given a complete graph dataset $\mathcal{G}$, we first decompose it into $K$ latent tasks:
\[
\mathcal{D}_1, \mathcal{D}_2, \dots, \mathcal{D}_K.
\]
Each task corresponds to a subset of nodes (or samples) defined according to standard class-incremental splits. These task identities are \emph{not exposed} during training, and are only used to construct the underlying data distributions.

\subsection{Temporal Scheduling}

We model the data stream at the mini-batch level, where time is indexed by batch order. Each task is assigned a center $\mu_k$ on the time axis to define its temporal position in the stream. Let $N_k$ denote the number of nodes in task $k$, and $B_k = \lceil N_k / B \rceil$ be the number of batches given batch size $B$. We define:
\begin{equation}
\mu_k = \sum_{i<k} B_i + \frac{B_k}{2}.
\end{equation}
Under a conventional sequential setting, this construction places each task in a contiguous time window. In our framework, it serves as a reference schedule around which transition dynamics are defined. Please find the detailed description in Appendix~\ref{app:data_construction}

\begin{table}[t]
\centering
\caption{Structural properties of benchmark datasets. Homophily $h$ measures the fraction of edges connecting nodes with the same label.}
\label{tab:dataset_properties}
\small
\begin{tabular}{lcccc}
\toprule
Property & CoraFull-CL~\cite{andrew2000cora} & Arxiv-CL~\cite{hu2020open} & Reddit-CL~\cite{hamilton2017sage} & RomanEmpire-CL~\cite{platonov2023roman} \\
\midrule
% Data source & CoraFull \cite{andrew2000cora} & OGB-Arxiv \cite{hu2020open} & Reddit \cite{hamilton2017sage} & RomanEmpire \cite{platonov2023roman} \\
Homophily $h$ & 0.57 & 0.66 & 0.76 & 0.05 \\
Avg.\ degree  & 13.2 & 13.8 & 1007.4 & 2.9 \\
Density ($\times 10^{-3}$) & 0.67 & 0.08 & 4.42 & 0.13 \\
\# nodes & 19,793 & 169,343 & 227,853 & 22,662 \\
\# edges & 130,622 & 1,166,243 & 114,615,892 & 32,927 \\
\# classes & 70 & 40 & 40 & 18 \\
\# tasks & 35 & 20 & 20 & 9 \\
\bottomrule
\end{tabular}
\vspace{-2mm}
\end{table}

\subsection{A Gaussian Parameterization of Transitions}

We instantiate the mixing function $\alpha_k(t)$ using a Gaussian kernel centered at $\mu_k$:
\begin{equation}
\alpha_k(t) = 
\frac{\exp\left(-\frac{(t - \mu_k)^2}{2\sigma^2}\right)}
{\sum_j \exp\left(-\frac{(t - \mu_j)^2}{2\sigma^2}\right)}.
\end{equation}
This choice is not restrictive; rather, it provides a simple and continuous parameterization that captures a wide range of transition patterns within a single framework. The parameter $\sigma$ controls the degree of temporal overlap across tasks. Small $\sigma$ produces near-discrete transitions, while larger $\sigma$ induces smooth and globally mixed distributions.

\subsection{Batch-Level Sampling}
\label{sec:batch_samp}
At each batch $t$, we construct a mini-batch by sampling from all tasks according to $\alpha_k(t)$.
Specifically, given batch size $B$, we compute:
\begin{equation}
    n_k(t) = \lfloor \alpha_k(t) \cdot B \rfloor,
\end{equation}
and allocate the remaining samples using largest-remainder rounding to ensure $\sum_k n_k(t) = B$. 

For each task $k$, we sample $n_k(t)$ instances from $\mathcal{D}_k$. This procedure provides a discrete approximation to the continuous mixture distribution defined by $\alpha_k(t)$.

We consider two sampling strategies:

\textbf{Without-replacement sampling.}
We maintain a per-task shuffled buffer and draw samples without replacement within each epoch. Once exhausted, the buffer is reshuffled. This mimics standard stochastic gradient descent behavior.

\textbf{With-replacement sampling.}
Samples are drawn independently with replacement according to the task distribution.

Finally, all sampled instances are shuffled within the batch to remove ordering bias.

\subsection{Transition Spectrum}

The Gaussian parameterization provides a unified view of several commonly used transition settings. As $\sigma \rightarrow 0$, the distribution approaches a piecewise constant schedule, recovering the conventional hard task transition setting. For small but non-zero $\sigma$, adjacent tasks overlap only within a limited temporal neighborhood, producing locally blurred boundaries. As $\sigma$ increases, all tasks maintain non-zero probability throughout the stream, resulting in globally mixed distributions. Intermediate values of $\sigma$ induce smooth and continuous distribution drift.This unified parameterization allows us to study the effect of transition dynamics within a single framework, without introducing separate benchmark designs for each setting.

\paragraph{Relation to standard protocols.} While the Gaussian formulation captures these settings as limiting cases, some protocols (e.g., hard transitions) cannot be exactly realized due to finite batch sizes. Therefore, for completeness, we also evaluate methods under the original settings and discuss the impact of batch size in Appendix~\ref{app:res}.
\vspace{-1mm}
\paragraph{Measuring Task Overlap.} To quantify the degree of task overlap, we define an overlap index:
\begin{equation}
\mathcal{O} = \frac{1}{T} \sum_{t=1}^{T} 
\mathbb{I}\left(\max_k \alpha_k(t) < \tau \right),
\end{equation}
where $\tau$ is a dominance threshold. This measures the fraction of time steps where no single task dominates, providing a quantitative measure of distribution smoothness.

\vspace{-2mm}
\label{sec:experiments}

\begin{table}[t]
  \centering
  \caption{Performance under continuous transition dynamics induced by Gaussian mixing curves. Metrics are mean\,$\pm$\,std over 3 runs ($\uparrow$ higher is better). Top-2 results per column are marked \textbf{1st}~/~\underline{2nd}.}
  \label{tab:tfogaussian}
  \setlength{\tabcolsep}{1mm}
  \scalebox{0.8}{
  \begin{tabular}{lcccccccc}
    \toprule
    \multirow{2}{*}{Method}
      & \multicolumn{2}{c}{Arxiv-CL}
      & \multicolumn{2}{c}{CoraFull-CL}
      & \multicolumn{2}{c}{Reddit-CL}
      & \multicolumn{2}{c}{RomanEmpire-CL}\\
    \cmidrule(lr){2-3}\cmidrule(lr){4-5}\cmidrule(lr){6-7}\cmidrule(l){8-9}
      & $A_{\text{AUC}}$\,\%\,$\uparrow$ & $\text{AF}_{\text{s}}$\,\%\,$\uparrow$
      & $A_{\text{AUC}}$\,\%\,$\uparrow$ & $\text{AF}_{\text{s}}$\,\%\,$\uparrow$
      & $A_{\text{AUC}}$\,\%\,$\uparrow$ & $\text{AF}_{\text{s}}$\,\%\,$\uparrow$
      & $A_{\text{AUC}}$\,\%\,$\uparrow$ & $\text{AF}_{\text{s}}$\,\%\,$\uparrow$ \\
    \midrule
    Bare
      & \vN{18.5}{1.5} & \vN{-65.4}{3.1}
      & \vN{21.9}{0.8} & \vN{-60.5}{7.3}
      & \vN{35.3}{4.2} & \vN{-35.0}{6.5}
      & \vN{39.0}{0.6} & \vN{-28.8}{3.6} \\
    Joint (Offline)
      & \vN{71.6}{1.4} & -
      & \vN{86.3}{0.1} & -
      & \vN{98.0}{0.1} & -
      & \vN{69.2}{1.1} & - \\
    \midrule
    A-GEM~\cite{chaudhry2018efficient}
      & \vN{34.1}{1.4} & \vN{-48.6}{4.4}
      & \vB{29.9}{2.8} & \vN{-53.9}{4.8}
      & \vN{34.4}{0.9} & \vN{-53.1}{9.1}
      & \vA{47.1}{0.5} & \vN{-21.7}{3.8} \\
    ER~\cite{chaudhry2019rs}
      & \vB{34.9}{0.8} & \vN{-37.3}{2.8}
      & \vN{27.8}{0.6} & \vN{-48.0}{5.2}
      & \vA{39.9}{0.2} & \vA{-40.6}{2.9}
      & \vN{43.6}{0.4} & \vN{-17.5}{3.0} \\
    GSS~\cite{aljundi2019gradient}
      & \vN{24.2}{3.2} & \vN{-60.4}{5.6}
      & \vN{29.3}{1.4} & \vN{-55.3}{0.1}
      & \vN{27.9}{2.0} & \vN{-55.5}{1.0}
      & \vB{46.4}{1.0} & \vB{-13.2}{2.2} \\
    MAS$^*$~\cite{aljundi2019task}
      & \vA{38.4}{2.1} & \vA{-22.2}{1.8}
      & \vN{29.8}{2.2} & \vB{-44.8}{8.0}
      & \vN{21.6}{2.4} & \vN{-83.5}{1.2}
      & \vN{43.0}{1.7} & \vN{-16.3}{2.1} \\
    SSM~\cite{zhang2022sparsified}
      & \vN{28.0}{4.3} & \vN{-64.5}{1.9}
      & \vN{25.4}{2.2} & \vN{-58.4}{3.8}
      & \vN{38.0}{1.6} & \vB{-45.3}{8.9}
      & \vN{42.2}{1.0} & \vN{-23.1}{4.1} \\
    SEM~\cite{zhang2023ricci}
      & \vN{26.8}{1.0} & \vN{-59.7}{6.9}
      & \vN{27.5}{1.9} & \vN{-58.4}{2.6}
      & \vN{34.4}{3.8} & \vN{-47.0}{7.8}
      & \vN{42.8}{2.0} & \vN{-29.0}{6.5} \\
    DMSG~\cite{qiao2025towards}
      & \vN{31.6}{0.9} & \vB{-31.9}{2.9}
      & \vA{34.1}{1.0} & \vA{-37.6}{3.5}
      & \vB{39.2}{1.2} & \vN{-47.0}{7.3}
      & \vN{43.8}{1.3} & \vA{-7.2}{4.2} \\
    \bottomrule
  \end{tabular}}
  \vspace{-2mm}
\end{table}

\begin{figure}[t]
    \centering
    \includegraphics[width=0.9\columnwidth]{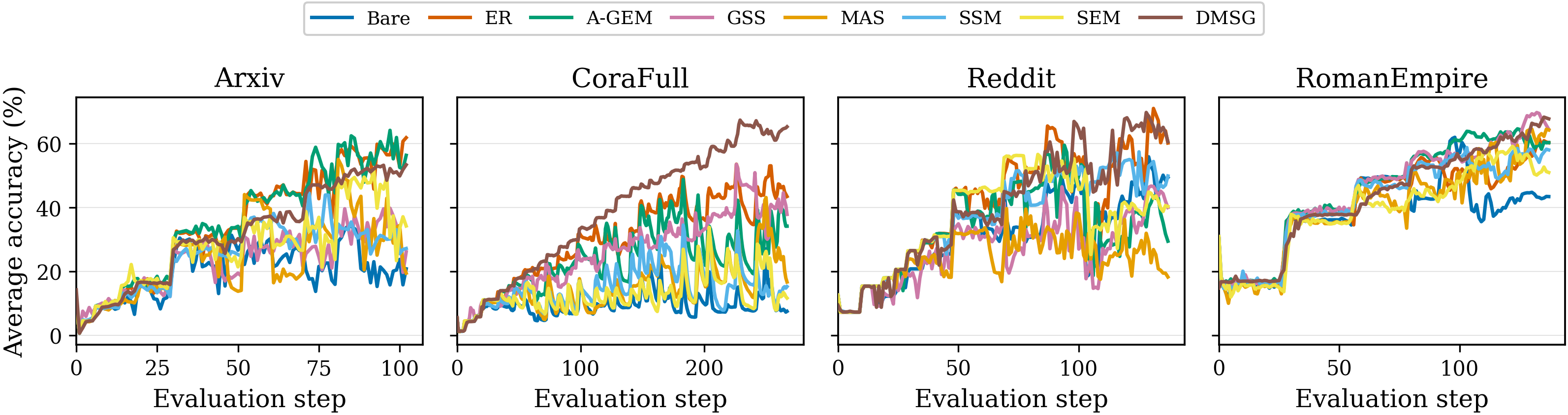}
    \vspace{-1mm}
    \caption{The dynamics of test accuracy of all implemented baselines on four datasets.}
    \vspace{-5mm}
    \label{fig:curve_gau}
\end{figure}

\begin{table}[t]
  \centering
  \caption{Performance under hard transition dynamics. Each latent task introduces new classes without overlap between consecutive segments of the stream. Mean\,$\pm$\,std over 3 runs ($\uparrow$ higher is better). }
  \label{tab:tfocis}
  \setlength{\tabcolsep}{1mm}
  \scalebox{0.8}{
  \begin{tabular}{lcccccccc}
    \toprule
    \multirow{2}{*}{Method}
      & \multicolumn{2}{c}{Arxiv-CL}
      & \multicolumn{2}{c}{CoraFull-CL}
      & \multicolumn{2}{c}{Reddit-CL}
      & \multicolumn{2}{c}{RomanEmpire-CL}\\
    \cmidrule(lr){2-3}\cmidrule(lr){4-5}\cmidrule(lr){6-7}\cmidrule(l){8-9}
      & AA\,\%\,$\uparrow$ & AF\,\%\,$\uparrow$
      & AA\,\%\,$\uparrow$ & AF\,\%\,$\uparrow$
      & AA\,\%\,$\uparrow$ & AF\,\%\,$\uparrow$
      & AA\,\%\,$\uparrow$ & AF\,\%\,$\uparrow$ \\
    \midrule
    Bare
      & \vN{12.2}{2.2} & \vN{-79.7}{0.9}
      & \vN{14.1}{1.3} & \vN{-75.3}{2.0}
      & \vN{28.3}{2.8} & \vN{-66.1}{8.1}
      & \vN{49.4}{2.1} & \vN{-40.9}{2.7} \\
    \midrule
    A-GEM~\cite{chaudhry2018efficient}
      & \vN{28.2}{1.6} & \vN{-65.3}{1.8}
      & \vN{28.6}{5.4} & \vN{-62.5}{5.3}
      & \vN{25.2}{7.1} & \vN{-70.3}{5.2}
      & \vN{49.9}{9.0} & \vN{-40.5}{9.0} \\
    ER~\cite{chaudhry2019rs}
      & \vB{47.5}{0.8} & \vN{-42.9}{0.9}
      & \vB{39.5}{5.9} & \vB{-52.3}{5.6}
      & \vA{55.9}{8.2} & \vA{-42.8}{8.1}
      & \vB{59.2}{1.4} & \vN{-29.1}{1.1} \\
    GSS~\cite{aljundi2019gradient}
      & \vN{28.7}{0.9} & \vN{-64.1}{0.9}
      & \vN{37.5}{3.4} & \vN{-55.0}{3.2}
      & \vB{55.6}{3.6} & \vB{-43.5}{3.4}
      & \vA{68.3}{1.5} & \vB{-20.7}{1.3} \\
    MAS$^*$~\cite{aljundi2019task}
      & \vA{55.5}{6.0} & \vB{-35.0}{5.9}
      & \vN{17.0}{1.1} & \vN{-73.3}{1.4}
      & \vN{49.9}{1.2} & \vN{-48.7}{1.5}
      & \vN{47.7}{7.1} & \vN{-42.4}{6.4} \\
    SSM~\cite{zhang2022sparsified}
      & \vN{27.5}{5.8} & \vN{-65.8}{5.9}
      & \vN{20.5}{3.1} & \vN{-69.8}{3.0}
      & \vN{40.9}{6.3} & \vN{-57.2}{6.1}
      & \vN{46.6}{5.9} & \vN{-43.9}{5.9} \\
    SEM~\cite{zhang2023ricci}
      & \vN{27.9}{1.6} & \vN{-65.4}{1.5}
      & \vN{16.6}{4.2} & \vN{-74.7}{4.9}
      & \vN{44.6}{9.2} & \vN{-53.6}{9.2}
      & \vN{47.6}{2.9} & \vN{-43.1}{3.2} \\
    DMSG~\cite{qiao2025towards}
      & \vN{42.0}{2.3} & \vA{-29.2}{2.3}
      & \vA{59.9}{1.6} & \vA{-21.7}{2.5}
      & \vN{28.4}{4.2} & \vN{-50.4}{7.7}
      & \vN{47.3}{3.8} & \vA{-10.0}{2.5} \\
    \bottomrule
  \end{tabular}}
  \vspace{-4mm}
\end{table}

\section{Experiments}
\label{sec:experiments}

\subsection{Experimental Setup}

\textbf{Datasets.}
We construct four benchmark datasets from widely used graph learning benchmarks. Each dataset is partitioned into a sequence of latent tasks using class-incremental splits, which define the underlying distributions but are never exposed during training. Table~\ref{tab:dataset_properties} summarizes the main graph statistics. The datasets cover both homophily-dominated graphs (e.g., Arxiv and Reddit) and heterophily settings (RomanEmpire~\cite{platonov2023roman}), reflecting the structural diversity commonly observed in real-world graph data~\cite{zhu2020beyond}. This diversity allows us to study how different transition dynamics interact with distinct relational structures and inductive biases.

\textbf{Baselines.}
We evaluate representative methods from three continual learning paradigms: rehearsal-based methods (ER~\cite{chaudhry2019rs}, A-GEM~\cite{chaudhry2018efficient}, GSS~\cite{aljundi2019gradient}), regularization-based methods (MAS*~\cite{aljundi2019task}), and graph-specific continual graph learning methods (SSM~\cite{zhang2022sparsified}, SEM~\cite{zhang2023ricci}, DMSG~\cite{qiao2025towards}). All methods are adapted to the task-free setting, where task identity is unavailable throughout training. We additionally include Bare (online training without continual learning strategies) as a lower bound and Joint (offline training on all data simultaneously) as an upper bound. Additional baseline details are provided in Appendix~\ref{app:de_datasets}.

\textbf{Implementation.}
All methods use a 2-layer GCN backbone~\cite{kipf2016semi} trained with Adam (learning rate $0.005$) in an online single-pass setting. Replay-based methods use a memory budget of 100 nodes. Each incoming batch is processed for one epoch before the next batch arrives. Results are averaged over 3 random seeds. Please see Appendix~\ref{app:backbone_res} for the discussion of the other two backbones.

\textbf{Evaluation protocol.}
We evaluate models continuously along the data stream. After processing each batch, the model is evaluated on the test sets associated with all latent tasks. For settings with hard transitions, we report the standard metrics used in continual learning: \textbf{Average Accuracy (AA)} measures final performance across all tasks: $\text{AA} = \frac{1}{K} \sum_{k=1}^{K} A_{B,k}$, where $A_{i,k}$ denotes the accuracy on the test set of latent task $k$ after observing the first $i$ batches, and $B$ is the total number of batches. \textbf{Average Forgetting (AF)} measures performance degradation: $\text{AF} = \frac{1}{K} \sum_{k=1}^{K} \left(\max_{i \le B} A_{i,k} - A_{B,k}\right)$.

Under smooth task-free transitions, explicit task boundaries are no longer available. Following~\cite{koh2022online}, we therefore report the \textbf{Area Under the Accuracy Curve (AUC)}, which measures performance throughout the stream:
\vspace{-2mm}
\begin{align}
A_{\text{AUC}} = \frac{1}{M} \sum_{j=1}^{M} \bar{A}(j \cdot \Delta),
\vspace{-2mm}
\end{align}
where $\bar{A}(t)$ denotes the average accuracy across all latent task test sets after observing $t$ batches, and $\Delta$ is the evaluation interval. To quantify forgetting without relying on explicit boundaries, we additionally report:
\vspace{-2mm}
\begin{align}
\text{AF}_{\text{s}} =
\frac{1}{K}
\sum_{k=1}^{K}
\left(
\max_{i \le B} A_{i,k}
-
A_{B,k}
\right).
\vspace{-2mm}
\end{align}

Unless otherwise specified, Gaussian-transition experiments use an intermediate overlap level (moderate $\sigma$), which balances task separation and distribution mixing and better reflects realistic non-stationary environments.

\subsection{Main Results under Task-Free Streaming}

Table~\ref{tab:tfogaussian} shows that existing continual graph learning methods degrade substantially under task-free streaming. Compared with the Joint upper bound, the best task-free method under Gaussian mixing reaches only $34.9\%$ vs.~$71.6\%$ on Arxiv-CL, $34.1\%$ vs.~$86.3\%$ on CoraFull-CL, and $39.9\%$ vs.~$98.0\%$ on Reddit-CL, leaving gaps of $35$--$60$ accuracy points. These results suggest that current continual graph learning methods remain far from the achievable upper bound once explicit task structure is removed.

The relative advantage over the Bare baseline also contracts substantially as transitions become smoother. On Arxiv-CL, the best continual learning method improves over Bare by $+43.3\%$ under hard transitions (MAS$^*$ in Table~\ref{tab:tfocis}), but only $+16.4\%$ under Gaussian mixing (ER in Table~\ref{tab:tfogaussian}), and further shrinks to $+5.5\%$ under $30\%$ global mixing (ER in Table~\ref{tab:blurry30}). A similar trend appears on RomanEmpire-CL, where the gap contracts from $+18.9\%$ under hard transitions (A-GEM in Table~\ref{tab:tfocis}) to $+8.1\%$ under Gaussian mixing (A-GEM in Table~\ref{tab:tfogaussian}).

This contraction persists even when the overall performance remains far from saturation. On Arxiv-CL under $30\%$ global mixing (Table~\ref{tab:blurry30}), Bare achieves only $22.6\%$ AA while Joint reaches $71.6\%$, yet the best continual learning method improves over Bare by merely $+5.5\%$. In contrast, on Reddit-CL the smaller gap is partially explained by a ceiling effect, since Bare already reaches $79.1\%$ AA under $30\%$ global mixing, leaving limited room for further improvement. Overall, these observations suggest that much of the benefit of existing continual learning methods originates from the task-partitioned structure implicitly provided by traditional task-based protocols. Once this structure becomes blurred, methods regress sharply toward the Bare baseline.

\subsection{Effect of Transition Dynamics}

\begin{wrapfigure}{r}{0.5\linewidth}
\vspace{-3mm}
    \centering
    \includegraphics[width=0.92\linewidth]{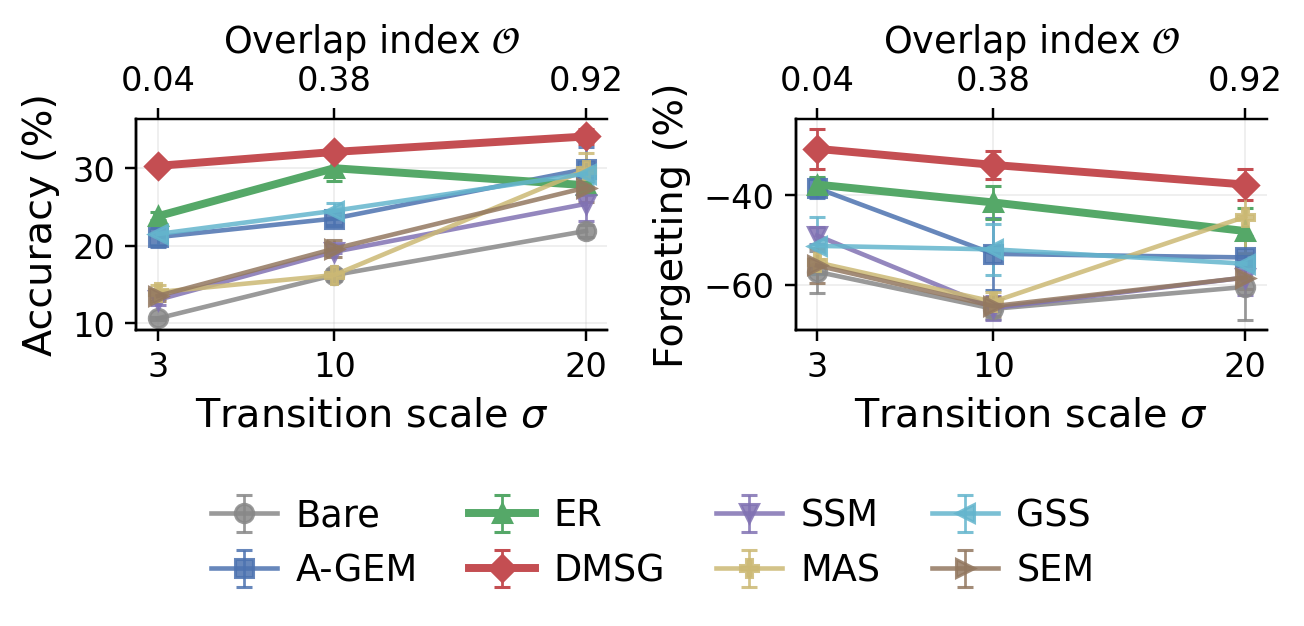}
    \vspace{-2mm}
    \caption{Effect of the transition scale $\sigma$ on CoraFull-CL.}
    \vspace{-3mm}
    \label{fig:sigma_sweep_cora}
\end{wrapfigure}

To study the influence of transition smoothness, we sweep the Gaussian mixing parameter $\sigma$ on CoraFull-CL across three regimes (Figure~\ref{fig:sigma_sweep_cora}): nearly disjoint transitions ($\sigma{=}3$, $\mathcal{O}{=}0.04$), moderate overlap ($\sigma{=}10$, $\mathcal{O}{=}0.38$), and heavy overlap ($\sigma{=}20$, $\mathcal{O}{=}0.92$). Numerical results are reported in Table~\ref{tab:cora_sigma_sweep}. As overlap increases, nearly all methods achieve higher $A_{\text{AUC}}$. For example, DMSG improves from $30.3\%$ at $\sigma{=}3$ to $34.1\%$ at $\sigma{=}20$, while Bare improves from $10.6\%$ to $21.9\%$. This suggests that mixed batches provide a form of implicit rehearsal, since each batch contains samples from multiple latent distributions rather than a single dominant task. 
However, improved adaptation is accompanied by increased forgetting. ER degrades from $-37.6\%$ forgetting at $\sigma{=}3$ to $-48.0\%$ at $\sigma{=}20$, while A-GEM drops from $-38.4\%$ to $-53.9\%$. Although smoother transitions improve online adaptation, they simultaneously weaken the effective training signal associated with each latent distribution, making old knowledge harder to preserve. This reveals a clear adaptation--forgetting trade-off under overlapping transitions. Among all baselines, DMSG exhibits the most stable forgetting behavior ($-29.6\% \rightarrow -33.2\% \rightarrow -37.6\%$), suggesting that its diversity-based sampling and VAE-based memory mechanism are more robust under continuously overlapping distributions.
\vspace{-2mm}
\subsection{Failure of Task-Based Assumptions}

Although our protocol does not expose task identity during training, several existing continual graph learning methods still implicitly rely on task-partitioned structure. Under mixing transitions, these assumptions become increasingly unreliable.

MAS$^*$ provides a representative example. The method relies on a loss-based heuristic to detect distribution shifts and trigger parameter regularization updates. Under hard transitions, where shift signals are sharp and localized, MAS$^*$ performs competitively (Table~\ref{tab:tfocis}). However, performance degrades substantially once transitions become smooth. On Arxiv-CL, MAS$^*$ drops from $55.5\%$ under hard transitions to $23.1\%$ under Gaussian mixing and $22.5\%$ under $30\%$ global mixing, approaching the Bare baseline ($22.6\%$). On Reddit-CL, it even falls below Bare under both Gaussian mixing ($21.6\%$ vs.~$35.3\%$) and $10\%$ global mixing ($45.0\%$ vs.~$52.8\%$). These results suggest that blurred transitions make the shift detector unreliable, leading to poorly timed parameter updates.

SSM and SEM exhibit a similar limitation. Both methods rely on task-incremental subgraph selection strategies, where task identity implicitly defines meaningful graph partitions. Under task-free mixing transitions, these partitions become ambiguous, substantially reducing the effectiveness of importance-based or curvature-based replay. Consequently, both methods consistently underperform the simpler ER baseline under Gaussian and global mixing settings. On CoraFull-CL under $30\%$ global mixing, ER achieves $49.0\%$ AA, while SSM and SEM reach only $20.2\%$ and $21.5\%$, respectively. Overall, these results suggest that many existing continual graph learning methods remain fundamentally tied to assumptions inherited from task-based continual learning.

% To further isolate the role of task identity, we additionally remove task information at inference time (Appendix~\ref{app:taskid_inf}, Table~\ref{tab:taskid_auc}). The resulting performance gap is small and occasionally reversed, indicating that the primary dependence on task identity arises during optimization rather than evaluation. 

\subsection{Additional Protocol Analysis}

\begin{wrapfigure}{r}{0.6\linewidth}
\vspace{-3mm}
    \centering
    \includegraphics[width=0.95\linewidth]{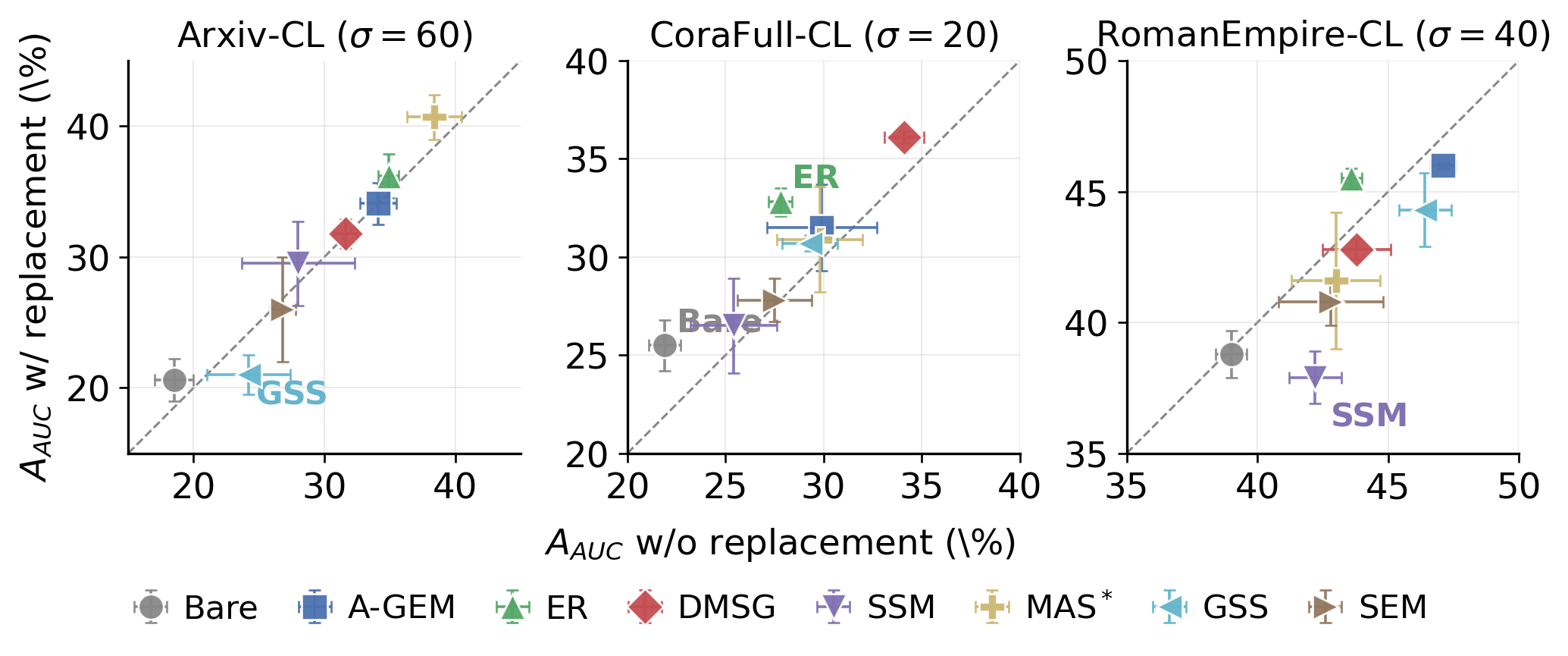}
    \vspace{-3mm}
    \caption{Effect of with- vs.\ without-replacement sampling on $A_{\text{AUC}}$ under Gaussian mixing.}
    \label{fig:rep_samp_comp}
\vspace{-3mm}
\end{wrapfigure}

\paragraph{Effect of with-replacement sampling.}
Figure~\ref{fig:rep_samp_comp} compares with- and without-replacement sampling strategies. On Arxiv-CL and CoraFull-CL, most methods lie above the diagonal, indicating that with-replacement sampling generally improves $A_{\text{AUC}}$. Repeated exposure to dominant-task samples introduces an additional form of implicit rehearsal, which stabilizes learning and improves adaptation.
% However, this trend does not hold on RomanEmpire-CL, where most methods fall below the diagonal. In heterophily-dominated settings with stronger structural diversity, repeated sampling may reduce effective batch diversity and negatively affect learning. Despite these differences, the relative ordering of methods remains largely stable across sampling strategies. These observations suggest that the interaction between sampling strategy and graph structure is non-trivial and should be considered when designing realistic task-free continual graph learning benchmarks.
However, this trend reverses on RomanEmpire-CL, where most methods fall below the diagonal. Unlike homophily-dominated graphs, heterophily graphs rely more heavily on structurally diverse neighborhoods for representation learning. Repeated sampling therefore increases local redundancy while reducing exposure to distinct neighborhood patterns across the stream. In the online setting, this can amplify noisy or incompatible message-passing signals, leading to less stable optimization despite the additional implicit rehearsal. These observations suggest that the interaction between sampling strategy and graph structure is non-trivial and should be considered when designing realistic task-free continual graph learning benchmarks.

\vspace{-2mm}
\section{Conclusion}
\label{sec:conclusion}
\vspace{-2mm}

We introduced \emph{DRIFT}, a benchmark for continual graph learning that replaces discrete task boundaries with continuously evolving latent task mixtures. By parameterizing transition dynamics through Gaussian mixing curves, DRIFT provides a unified framework spanning a spectrum of non-stationary settings, from nearly disjoint transitions to heavily overlapping streams. This formulation enables a systematic study of how transition smoothness influences continual graph learning behavior. 

Across four datasets and multiple representative baselines, we observe that existing continual graph learning methods degrade substantially once explicit task structure becomes blurred. In particular, the relative advantage of continual learning methods over the ``Bare”  baseline contracts consistently as overlap increases, even in settings where performance remains far from the ``Joint” upper bound. These findings suggest that many current methods derive a significant portion of their effectiveness from assumptions implicitly tied to sharp task boundaries. 

Our analysis further reveals that transition smoothness introduces a non-trivial adaptation--forgetting trade-off. While overlapping batches provide implicit rehearsal and improve online adaptation, they simultaneously weaken the separability of latent distributions, making old knowledge more difficult to preserve. Methods designed around explicit task partitioning, such as shift-triggered regularization or task-aware replay, become particularly unstable under such conditions. 

Overall, our results indicate that future continual graph learning research should move beyond from boundary-dependent formulations to techniques that explicitly account for continuous distribution drift and overlapping task structure, and push the frontier to open-world graph settings.

\vspace{-3mm}
\section{Limitations and Future Work}
\label{sec:limitations}

This work focuses primarily on the role of transition dynamics in continual graph learning and therefore has several limitations. First, our evaluation is restricted to node classification. Extending the benchmark to other graph learning settings, such as link prediction, graph classification, and temporal reasoning, would provide a more comprehensive evaluation of task-free continual graph learning. Second, DRIFT models non-stationarity using Gaussian transition dynamics. Although this provides a controllable and interpretable framework for studying overlap, real-world environments may exhibit more complex forms of drift, including abrupt distribution shifts, periodic transitions, long-range recurrence, or evolving graph topology. Incorporating these transition patterns is an important direction for future work. Third, the current benchmark is constructed from static graph datasets with synthetic streaming protocols. While this allows controlled analysis of transition dynamics, future extensions could incorporate naturally evolving graph streams collected from real-world systems such as social networks, recommendation platforms, or dynamic knowledge graphs. Finally, our findings suggest that many existing continual graph learning methods remain fundamentally tied to assumptions inherited from task-based continual learning. Developing methods that operate without explicit task segmentation, while remaining robust to continuous and overlapping distribution shifts, remains an open challenge for future research.

\section{Acknowledgment}
This work was supported by the National Science Foundation under Grant No. 2338878.

%reference
\bibliographystyle{unsrt}
\bibliography{ref}

%%%%%%%%%%%%%%%%%%%%%%%%%%%%%%%%%%%%%%%%%%%%%%%%%%%%%%%%%%%%

\appendix
% \section{Technical Appendices and Supplementary Material}
% Technical appendices with additional results, figures, graphs and proofs may be submitted with the paper submission before the full submission deadline (see above), or as a separate PDF in the ZIP file below before the supplementary material deadline. There is no page limit for the technical appendices.
\newpage
\section{Details of DRIFT Benchmark}
\label{app:details}
\subsection{Details of Benchmark Baselines}
\label{app:de_bl}
A brief introduction of the implemented Continual Learning (CL) methods is as below:
\begin{enumerate}
    \item \textbf{Bare model} denotes the backbone GNN without the continual learning technique. Therefore, this can be viewed as the lower bound on the continual learning performance.
    \item \textbf{A-GEM \cite{chaudhry2018efficient}} is an efficient version of GEM~\cite{lopez2017gradient}, which ensures that the average loss for historical tasks does not increase by projecting the gradient of incoming data onto the orthogonal space of the gradient of historical data. We use Reservoir Sampling to select nodes.
    \item \textbf{Experience Replay (ER) \cite{chaudhry2019rs}} selects nodes from the incoming batch to be stored in the memory buffer by Reservoir Sampling, which is a simple yet effective method for CL. New incoming batches for training are then augmented with nodes sampled uniformly from the buffer.
    \item \textbf{Gradient-based Sample Selection (GSS) \cite{aljundi2019gradient}} selects representative samples from the incoming data stream by measuring the diversity of their gradients. Specifically, it maintains samples whose gradients are less aligned with those already stored in the memory buffer, thereby promoting gradient diversity and reducing redundancy. New batches for training are also combined with samples uniformly from the buffer.
    \item \textbf{Memory Aware Synapses (MAS)* \cite{aljundi2019task}} is a task-free version of MAS~\cite{aljundi2018memory}, which adds a detector guiding the model when to update the important weights in a streaming fashion.
    \item \textbf{Sparsified Subgraph Memory (SSM) \cite{zhang2022sparsified}} stores representative subgraphs instead of individual nodes to preserve both structural and feature information. It constructs sparsified subgraphs by selecting important nodes based on their contribution to the graph topology to reduce redundancy. Reservoir sampling is used as the sampling strategy.
    \item \textbf{Subgraph Episodic Memory (SEM) \cite{zhang2023ricci}} extends subgraph-based memory by introducing a curvature-guided sparsification mechanism. It constructs Subgraph Episodic Memory (SEM) to store computation subgraphs, and further prunes edges based on Ricci curvature to preserve the most informative topological relationships for message passing. This approach reduces redundancy while maintaining critical structural information, enabling more efficient and effective memory replay in continual graph learning. 
    \item \textbf{Diversified Memory Selection and Generation (DMSG) \cite{qiao2025towards}} maintains a diversified memory buffer by jointly considering intra- and inter-class diversity when selecting samples. To adequately reuse the knowledge preserved in the buffer, it utilizes a variational layer to generate the distribution of buffer node embeddings and sample synthesized ones for replaying. To further enhance the quality of generated samples, it incorporates adversarial variational embedding learning and a reconstruction-based decoder to improve generalization.
\end{enumerate}

\subsection{Details of Benchmark Datasets}
\label{app:de_datasets}
\paragraph{CoraFull \cite{andrew2000cora}} The CoraFull contains 19,793 scientific publications classified into seventy classes, with a citation network of 130,622 links. Each publication is represented by a 1,433-dimensional binary word vector indicating the presence or absence of 1,433 unique dictionary words.
\paragraph{OGB-Arxiv \cite{hu2020open}} The ogbn-arxiv dataset is a directed graph representing the citation network of 169,343 Computer Science arXiv papers indexed by MAG, with 2,315,598 directed edges indicating citation relationships. Each paper is characterized by a 128-dimensional feature vector derived from averaging word embeddings of its title and abstract, where embeddings are trained via the skip-gram model on the MAG corpus. The prediction task involves classifying papers into 40 manually labeled arXiv CS subject areas (e.g., cs.AI, cs.LG), formulated as a 40-class classification problem to automate scientific publication topic categorization.
\paragraph{Reddit \cite{hamilton2017sage}} The Reddit dataset consists of posts from different communities of the Reddit platform, where nodes represent posts, and edges connect posts commented on by the same user, forming a large interaction graph.
\paragraph{RomanEmpire \cite{platonov2023roman}} The RomanEmpire dataset is constructed from the English Wikipedia article "Roman Empire" (one of the longest Wikipedia articles), featuring 22,662 nodes, 65,854 edges, 300-dimensional fastText word embeddings as node features, and 18 labeled syntactic roles.

\subsection{Data Stream Construction}
\label{app:data_construction}

We construct all task-free streams from static graph datasets by defining a sequence of latent task distributions and then generating mini-batches according to a time-dependent mixing schedule. Below we first describe the Gaussian protocol used in the main experiments, followed by the additional transition variants.

\paragraph{Latent task decomposition.}
Each dataset $\mathcal{G}$ is partitioned into $K$ latent tasks, with two classes assigned to each task. This gives $K{=}35$ for CoraFull-CL, $K{=}20$ for Arxiv-CL, $K{=}20$ for Reddit-CL, and $K{=}9$ for RomanEmpire-CL (Table~\ref{tab:dataset_properties}). Classes are grouped according to their original label ordering in the dataset. Let $\mathcal{D}_k$ denote the node set associated with latent task $k$. These task identities are used only to define the underlying distributions and are never provided to the model.

\paragraph{Gaussian transition schedule.}
For the Gaussian setting, each latent task $k$ is assigned a center $\mu_k$ along the stream:
\vspace{-2mm}
\begin{align}
\mu_k =
\sum_{i<k} \left\lceil \frac{N_i}{B} \right\rceil
+
\frac{1}{2}
\left\lceil \frac{N_k}{B} \right\rceil,
\vspace{-2mm}
\end{align}
where $N_k$ is the number of samples in $\mathcal{D}_k$ and $B$ is the mini-batch size. The total stream length is
$
T = \sum_k \lceil N_k / B \rceil.
$

At stream step $t$, the contribution of each latent task is determined by Gaussian weights
\vspace{-2mm}
\begin{align}
\alpha_k(t)
\propto
\exp\left(
-\frac{(t-\mu_k)^2}{2\sigma^2}
\right),
\vspace{-2mm}
\end{align}
followed by softmax normalization over all tasks. Larger $\sigma$ produces smoother transitions and stronger overlap between neighbouring distributions.

The main experiments use
$\sigma{=}20$ for CoraFull-CL,
$\sigma{=}60$ for Arxiv-CL,
$\sigma{=}120$ for Reddit-CL,
and $\sigma{=}40$ for RomanEmpire-CL.
These values place all datasets in the high-overlap regime. Additional results on CoraFull-CL further sweep $\sigma\in\{3,10,20\}$, corresponding to overlap indices $\mathcal{O}\in\{0.04,0.38,0.92\}$.

\paragraph{Mini-batch generation.}
At each step $t$, task $k$ contributes
$
n_k(t)=\lfloor \alpha_k(t)\cdot B \rfloor
$
samples to the current mini-batch. Remaining slots are assigned using largest-remainder rounding such that
$
\sum_k n_k(t)=B.
$

Unless otherwise stated, we use without-replacement sampling. Each latent task maintains an independently shuffled sample queue that is reshuffled once exhausted. Samples selected for the current step are merged and randomly permuted before being forwarded to the model. The default batch size is $B{=}10$ for all datasets.

\paragraph{Hard-transition variant.}
The hard-transition setting corresponds to the limiting case $\sigma\rightarrow0$, where each mini-batch is sampled entirely from a single latent task. Concretely, for
\vspace{-2mm}
\begin{align}
t \in
\left[
\mu_k-\left\lceil \frac{N_k}{2B}\right\rceil,
\,
\mu_k+\left\lceil \frac{N_k}{2B}\right\rceil
\right),
\vspace{-2mm}
\end{align}
all $B$ samples are drawn from $\mathcal{D}_k$. This recovers the standard class-incremental protocol commonly used in prior continual graph learning benchmarks.

\paragraph{Global-mixing variant.}
For global mixing, each latent task persistently contributes a fixed fraction $p$ of its samples throughout the stream. Specifically, we reserve
$
p\cdot |\mathcal{D}_k|
$
samples from each task as a shared pool injected into all stream regions, while the remaining samples follow the hard-transition schedule. We report results for $p{=}10\%$ and $p{=}30\%$.

\paragraph{Boundary-local mixing variant.}
For boundary-local mixing, overlap is introduced only near neighbouring task boundaries. We use a mixing window of $K_{\text{bl}}{=}5$ batches. The last
$
K_{\text{bl}}\cdot B
$
samples from task $k$ and the first
$
K_{\text{bl}}\cdot B
$
samples from task $k{+}1$
are pooled and redistributed uniformly across the corresponding boundary region. Outside these windows, the stream remains task-pure.

\paragraph{Reproducibility.}
For every $(\text{dataset}, \sigma, \text{sampling})$ configuration, the full stream is generated once and stored as a fixed sequence of mini-batches. All baselines therefore observe identical input streams under the same random seed. Stream construction and method-level randomness use separate random states initialized from the same seed.

\subsection{Training and Evaluation Configuration}
\label{app:training_config}

\paragraph{Model backbone.}
All methods use the same 2-layer GCN backbone~\cite{kipf2016semi} with hidden dimension $256$. We do not use dropout or batch normalization.

\paragraph{Optimization details.}
All experiments use Adam with learning rate $5\times10^{-3}$. The mini-batch size is fixed at $B{=}10$. Each incoming batch is processed for one epoch before the next batch arrives. No learning-rate scheduling, gradient clipping, or warm-up is applied.

\paragraph{Method-specific settings.}
Whenever possible, we follow the original hyperparameter settings reported in the corresponding papers and modify only parameters directly tied to the streaming setup.

For ER~\cite{chaudhry2019rs}, A-GEM~\cite{chaudhry2018efficient}, SSM~\cite{zhang2022sparsified}, SEM~\cite{zhang2023ricci}, and DMSG~\cite{qiao2025towards}, the replay memory size is fixed to $|\mathcal{M}|{=}100$ nodes. ER and A-GEM use reservoir sampling. SSM uses sparsified 2-hop replay subgraphs with fan-outs $(5,5)$, while SEM follows the original curvature-based pruning strategy. DMSG uses the original diversified variational replay mechanism. Replay samples are mixed with incoming samples using a $1{:}1$ ratio.

For GSS~\cite{aljundi2019gradient}, we use$n_{\text{sampled}}{=}100,$$n_{\text{constraints}}{=}10,$gradient memory strength $5$, and one inner iteration per step.

For MAS$^*$~\cite{aljundi2019task}, the regularization strength is set to $0.5$ (provided by original implementation code). The online detector updates parameter importance whenever the loss-difference criterion in the original implementation is triggered.

% \paragraph{Evaluation.}
% After each mini-batch, the current model is evaluated on the held-out test set of every latent task. Metrics reported in the main paper aggregate these online accuracies over the full stream. For AUC evaluation, we use $\Delta{=}1$, corresponding to evaluation after every batch.

\paragraph{Statistical reporting.}
All results are averaged over three runs with random seeds $\{1,2,3\}$. Reported uncertainties correspond to standard deviations across runs.
All experiments are conducted on a single NVIDIA A6000 GPU using PyTorch and DGL.

\section{Additional Results}
\label{app:res}

We provide the detailed experiment results and additional analysis in this section.

\subsection{Results on additional backbones.}
\label{app:backbone_res}
\begin{table*}[t]
\centering
\small
\setlength{\tabcolsep}{4pt}
\caption{Results under Gaussian mixing using different backbone architectures. We report $A_{\text{AUC}}$ ($\uparrow$) and $\mathrm{AF}_{\mathrm{s}}$ ($\uparrow$).}
\label{tab:backbone_results}
\setlength{\tabcolsep}{1mm}
\scalebox{1.0}{
\begin{tabular}{clcccccc}
\toprule
\multirow{2}{*}{Dataset} & \multirow{2}{*}{Method}
& \multicolumn{2}{c}{GAT}
& \multicolumn{2}{c}{GIN} \\
\cmidrule(lr){3-4} \cmidrule(lr){5-6}
& & $A_{\text{AUC}}$ & $\mathrm{AF}_{\mathrm{s}}$
& $A_{\text{AUC}}$ & $\mathrm{AF}_{\mathrm{s}}$ \\
\midrule

\multirow{7}{*}{Arxiv-CL}
& Bare  & $27.4_{\pm1.9}$ & $-54.6_{\pm1.6}$ & $18.5_{\pm0.3}$ & $-68.2_{\pm5.0}$ \\
& A-GEM~\cite{chaudhry2018efficient} & $32.1_{\pm0.6}$ & $-48.8_{\pm2.3}$ & $32.6_{\pm0.5}$ & $-53.5_{\pm8.7}$ \\
& ER~\cite{chaudhry2019rs}    & $\mathbf{33.2_{\pm1.6}}$ & $-52.2_{\pm7.4}$ & $34.8_{\pm0.9}$ & $-40.7_{\pm1.2}$ \\
& SSM~\cite{zhang2022sparsified}   & $23.2_{\pm5.2}$ & $-63.5_{\pm2.0}$ & $27.7_{\pm1.2}$ & $-62.6_{\pm4.5}$ \\
& MAS$^*$~\cite{aljundi2019task}   & $27.4_{\pm1.9}$ & $-54.6_{\pm1.6}$ & $\mathbf{39.7_{\pm4.9}}$ & $\mathbf{-18.9_{\pm3.5}}$ \\
& GSS~\cite{aljundi2019gradient}   & $22.1_{\pm1.7}$ & $-64.2_{\pm8.2}$ & $23.1_{\pm0.3}$ & $-64.1_{\pm1.4}$ \\
& SEM~\cite{zhang2023ricci}   & $27.7_{\pm5.7}$ & $-63.4_{\pm3.5}$ & $23.8_{\pm2.0}$ & $-59.1_{\pm5.0}$ \\
\midrule

\multirow{7}{*}{CoraFull-CL}
& Bare  & $39.2_{\pm2.8}$ & $-38.4_{\pm14.0}$ & $27.2_{\pm1.1}$ & $-56.2_{\pm7.6}$ \\
& A-GEM & $39.8_{\pm1.1}$ & $-31.6_{\pm6.5}$ & $32.3_{\pm0.9}$ & $-46.4_{\pm0.9}$ \\
& ER~\cite{chaudhry2019rs}    & $\mathbf{41.2_{\pm0.4}}$ & $\mathbf{-28.2_{\pm6.1}}$ & $\mathbf{32.8_{\pm1.8}}$ & $-44.4_{\pm3.0}$ \\
& SSM~\cite{zhang2022sparsified}   & $34.5_{\pm3.0}$ & $-37.3_{\pm11.0}$ & $25.7_{\pm1.4}$ & $-63.7_{\pm3.3}$ \\
& MAS$^*$~\cite{aljundi2019task}   & $39.2_{\pm2.8}$ & $-38.4_{\pm14.0}$ & $28.3_{\pm3.8}$ & $-47.0_{\pm10.4}$ \\
& GSS~\cite{aljundi2019gradient}   & $36.8_{\pm1.4}$ & $-38.3_{\pm8.9}$ & $30.2_{\pm1.1}$ & $\mathbf{-43.7_{\pm3.1}}$ \\
& SEM~\cite{zhang2023ricci}   & $35.6_{\pm1.4}$ & $-39.0_{\pm2.2}$ & $27.4_{\pm1.6}$ & $-51.0_{\pm1.5}$ \\
\midrule

\multirow{7}{*}{RomanEmpire-CL}
& Bare  & $44.1_{\pm0.5}$ & $-23.7_{\pm6.5}$ & $43.2_{\pm1.4}$ & $-18.7_{\pm1.6}$ \\
& A-GEM~\cite{chaudhry2018efficient} & $44.9_{\pm2.2}$ & $-22.2_{\pm5.9}$ & $46.7_{\pm0.2}$ & $-30.4_{\pm3.7}$ \\
& ER~\cite{chaudhry2019rs}    & $\mathbf{45.8_{\pm0.9}}$ & $-24.0_{\pm4.7}$ & $\mathbf{47.0_{\pm1.0}}$ & $-21.3_{\pm2.5}$ \\
& SSM~\cite{zhang2022sparsified}   & $38.8_{\pm4.0}$ & $-35.1_{\pm3.9}$ & $40.4_{\pm3.3}$ & $-27.6_{\pm7.6}$ \\
& MAS$^*$~\cite{aljundi2019task}   & $44.2_{\pm0.4}$ & $\mathbf{-21.6_{\pm3.7}}$ & $45.9_{\pm1.5}$ & $-18.3_{\pm6.7}$ \\
& GSS~\cite{aljundi2019gradient}   & $43.6_{\pm1.4}$ & $-21.7_{\pm2.2}$ & $45.1_{\pm1.5}$ & $\mathbf{-17.1_{\pm1.4}}$ \\
& SEM~\cite{zhang2023ricci}   & $35.6_{\pm4.4}$ & $-37.3_{\pm1.1}$ & $40.9_{\pm0.9}$ & $-25.5_{\pm1.4}$ \\
\bottomrule
\end{tabular}
}
\end{table*}

Table~\ref{tab:backbone_results} reports additional experiments using GAT and GIN backbones under Gaussian mixing. Overall, the main observations from the GCN setting remain consistent across architectures. In particular, replay-based methods such as ER and A-GEM remain comparatively stable across most datasets, while methods relying on explicit task structure (e.g., SSM and SEM) continue to degrade substantially under smooth transitions. This suggests that the failure modes observed in the main paper are not specific to a particular backbone architecture.

On Arxiv-CL and CoraFull-CL, replay-based methods consistently outperform Bare under both GAT and GIN. For example, under the GAT backbone, ER improves over Bare from $27.4\%$ to $33.2\%$ on Arxiv-CL and from $39.2\%$ to $41.2\%$ on CoraFull-CL. Similar trends hold under GIN. In contrast, SSM and SEM often perform close to or below Bare, particularly on CoraFull-CL under GIN, where SSM reaches only $25.7\%$ compared to $27.2\%$ for Bare. This is consistent with the observations in the main experiments that task-aware replay strategies become unreliable once task boundaries are blurred.

We also observe that the relative behaviour across methods is more stable than the absolute ranking of individual methods. For instance, MAS performs competitively on Arxiv-CL under the GIN backbone ($39.7\%$ AUC), but its behaviour remains inconsistent across datasets and transition regimes. Similarly, although GIN generally changes the absolute scale of performance compared to GCN and GAT, the overall trends regarding overlap sensitivity and boundary dependence remain unchanged.

Overall, these results suggest that the conclusions of the main paper are robust across different message-passing architectures and are primarily driven by the transition structure of the stream rather than by a specific backbone choice.

\begin{table}[htpb]
  \centering
  \caption{Numerical values for the $\sigma$ sweep on CoraFull-CL (companion to Figure~\ref{fig:sigma_sweep_cora}). Larger $\sigma$ produces stronger temporal overlap between latent task distributions, quantified by the overlap index $\mathcal{O}$. Mean$\,\pm\,$std over 3 runs ($\uparrow$ higher is better). Top-2 per column are marked \textbf{1st}~/~\underline{2nd}.}
  \label{tab:cora_sigma_sweep}
  \setlength{\tabcolsep}{1mm}
  \scalebox{0.9}{
  \begin{tabular}{lcccccc}
    \toprule
    \multirow{2}{*}{Method}
      & \multicolumn{2}{c}{$\sigma{=}3,\ \mathcal{O}{=}0.04$}
      & \multicolumn{2}{c}{$\sigma{=}10,\ \mathcal{O}{=}0.38$}
      & \multicolumn{2}{c}{$\sigma{=}20,\ \mathcal{O}{=}0.92$} \\
    \cmidrule(lr){2-3}\cmidrule(lr){4-5}\cmidrule(l){6-7}
      & $A_{\text{AUC}}$\,$\uparrow$ & $\text{AF}_{\text{s}}$\,$\uparrow$
      & $A_{\text{AUC}}$\,$\uparrow$ & $\text{AF}_{\text{s}}$\,$\uparrow$
      & $A_{\text{AUC}}$\,$\uparrow$ & $\text{AF}_{\text{s}}$\,$\uparrow$ \\
    \midrule
    Bare
      & \vN{10.6}{0.2}  & \vN{-57.1}{4.7}
      & \vN{16.2}{0.8}  & \vN{-65.4}{2.2}
      & \vN{21.9}{0.8}  & \vN{-60.5}{7.3} \\
    A-GEM~\cite{chaudhry2018efficient}
      & \vN{21.1}{1.3}  & \vN{-38.4}{2.2}
      & \vN{23.5}{0.8}  & \vN{-53.1}{8.0}
      & \vB{29.9}{2.8}  & \vN{-53.9}{4.8} \\
    ER~\cite{chaudhry2019rs}
      & \vB{23.8}{0.5}  & \vB{-37.6}{1.8}
      & \vB{30.0}{1.7}  & \vB{-41.6}{3.6}
      & \vN{27.8}{0.6}  & \vN{-48.0}{5.2} \\
    GSS~\cite{aljundi2019gradient}
      & \vN{21.5}{1.4}  & \vN{-51.3}{6.4}
      & \vN{24.5}{1.0}  & \vN{-52.1}{5.7}
      & \vN{29.3}{1.4}  & \vN{-55.3}{0.1} \\
    MAS$^*$~\cite{aljundi2019task}
      & \vN{14.0}{0.9}  & \vN{-55.0}{0.7}
      & \vN{16.2}{0.4}  & \vN{-63.8}{2.1}
      & \vN{29.8}{2.2}  & \vB{-44.8}{8.0} \\
    SSM~\cite{zhang2022sparsified}
      & \vN{13.0}{0.6}  & \vN{-49.0}{2.0}
      & \vN{19.2}{0.2}  & \vN{-65.2}{2.7}
      & \vN{25.4}{2.2}  & \vN{-58.4}{3.8} \\
    SEM~\cite{zhang2023ricci}
      & \vN{13.5}{0.7}  & \vN{-55.7}{3.9}
      & \vN{19.6}{1.1}  & \vN{-64.8}{2.1}
      & \vN{27.5}{1.9}  & \vN{-58.4}{2.6} \\
    DMSG~\cite{qiao2025towards}
      & \vA{30.3}{0.3}  & \vA{-29.6}{4.5}
      & \vA{32.1}{0.4}  & \vA{-33.2}{3.1}
      & \vA{34.1}{1.0}  & \vA{-37.6}{3.5} \\
    \bottomrule
  \end{tabular}}
\end{table}

\begin{table}[htpb]
  \centering
  \caption{Numerical values for the $\sigma$ sweep on RomanEmpire-CL. Larger $\sigma$ produces stronger temporal overlap between latent task distributions, quantified by the overlap index $\mathcal{O}$. Mean$\,\pm\,$std over 3 runs ($\uparrow$ higher is better). Top-2 per column are marked \textbf{1st}~/~\underline{2nd}.}
  \label{tab:roman_sigma_sweep}
  \setlength{\tabcolsep}{1mm}
  \scalebox{0.9}{
  \begin{tabular}{lcccccc}
    \toprule
    \multirow{2}{*}{Method}
      & \multicolumn{2}{c}{$\sigma{=}20,\ \mathcal{O}{=}0.11$}
      & \multicolumn{2}{c}{$\sigma{=}40,\ \mathcal{O}{=}0.31$}
      & \multicolumn{2}{c}{$\sigma{=}120,\ \mathcal{O}{=}0.88$} \\
    \cmidrule(lr){2-3}\cmidrule(lr){4-5}\cmidrule(l){6-7}
      & $A_{\text{AUC}}$\,$\uparrow$ & $\text{AF}_{\text{s}}$\,$\uparrow$
      & $A_{\text{AUC}}$\,$\uparrow$ & $\text{AF}_{\text{s}}$\,$\uparrow$
      & $A_{\text{AUC}}$\,$\uparrow$ & $\text{AF}_{\text{s}}$\,$\uparrow$ \\
    \midrule
    Bare
      & \vN{36.7}{1.0}  & \vN{-36.3}{4.6}
      & \vN{39.0}{0.6}  & \vN{-28.8}{3.6}
      & \vN{49.1}{0.6}  & \vN{-8.0}{0.5} \\
    A-GEM~\cite{chaudhry2018efficient}
      & \vB{43.1}{0.6}  & \vN{-33.0}{5.4}
      & \vA{47.1}{0.5}  & \vN{-21.7}{3.8}
      & \vA{49.2}{0.5}  & \vN{-9.6}{1.5} \\
    ER~\cite{chaudhry2019rs}
      & \vA{44.2}{0.3}  & \vB{-23.1}{3.2}
      & \vN{43.6}{0.4}  & \vN{-17.5}{3.0}
      & \vN{48.6}{0.1}  & \vB{-6.4}{1.4} \\
    GSS~\cite{aljundi2019gradient}
      & \vN{42.2}{0.7}  & \vN{-21.4}{2.1}
      & \vB{46.4}{1.0}  & \vB{-13.2}{2.2}
      & \vB{49.0}{0.2}  & \vN{-9.7}{1.4} \\
    MAS$^*$~\cite{aljundi2019task}
      & \vN{35.5}{1.1}  & \vN{-37.1}{2.8}
      & \vN{43.0}{1.7}  & \vN{-16.3}{2.1}
      & \vN{47.8}{1.4}  & \vN{-12.4}{3.1} \\
    SSM~\cite{zhang2022sparsified}
      & \vN{35.3}{0.6}  & \vN{-39.0}{4.2}
      & \vN{42.2}{1.0}  & \vN{-23.1}{4.1}
      & \vN{47.2}{2.3}  & \vN{-10.2}{2.5} \\
    SEM~\cite{zhang2023ricci}
      & \vN{38.9}{2.0}  & \vN{-33.4}{4.5}
      & \vN{42.8}{2.0}  & \vN{-29.0}{6.5}
      & \vN{48.9}{1.3}  & \vN{-7.5}{1.2} \\
    DMSG~\cite{qiao2025towards}
      & \vN{42.0}{0.1}  & \vA{-14.6}{2.8}
      & \vN{43.8}{1.3}  & \vA{-7.2}{4.2}
      & \vN{44.7}{0.7}  & \vA{-4.8}{0.5} \\
    \bottomrule
  \end{tabular}}
\end{table}

\subsection{Extended Study on Transition scale}
\label{app:sigma_sweep}
To assess the generality of the observed trends, we extend the $\sigma$ sweep 
to the more heterophilous RomanEmpire-CL dataset.

% Table~\ref{tab:cora_sigma_sweep} reports the numerical values that accompany Figure~\ref{fig:sigma_sweep_cora} in the main text. Two trends supplement the discussion in RQ2 and RQ4. First, methods that lag well behind the replay-based baselines under low overlap close most of that gap as $\sigma$ grows: MAS$^*$ rises from $14.0$ to $29.8$ AUC ($\sigma{=}3 \rightarrow 20$), essentially matching A-GEM's $29.9$ at the same $\sigma$, and SSM/SEM nearly double their AUC over the same range ($13.0 \rightarrow 25.4$ and $13.5 \rightarrow 27.5$). Under heavy overlap the choice of CL mechanism matters less than the implicit rehearsal already supplied by the data stream. Second, DMSG keeps the absolute lead at every $\sigma$ but its margin over Bare contracts from $+19.7$ to $+12.2$ AUC, mirroring the diminishing-leverage pattern reported in RQ4 and confirming that even drift-aware machinery extracts less marginal value as overlap saturates the stream.

Table~\ref{tab:cora_sigma_sweep} reports the numerical values that accompany Figure~\ref{fig:sigma_sweep_cora} in the main text. As $\sigma$ increases, all methods exhibit substantial gains in $A_{\text{AUC}}$, accompanied by reduced forgetting. In particular, Bare improves from $36.7$ to $49.1$ AUC ($\sigma{=}20 \rightarrow 120$), while its AF increases from $-36.3$ to $-8.0$, indicating that strong overlap effectively mitigates forgetting. This confirms that increasing overlap provides implicit rehearsal, consistent with the trends observed on CoraFull-CL.

However, unlike CoraFull-CL, the relative advantage of specialized CL methods diminishes more sharply under high overlap~\ref{tab:roman_sigma_sweep}. At $\sigma{=}120$ (overlap index $\mathcal{O}=0.88$), most methods converge to a narrow performance range ($\sim48$--$49$ AUC), with Bare already matching or exceeding several CL methods. This suggests a stronger ceiling effect, where the benefit of explicit CL mechanisms becomes marginal once the stream is highly mixed. In contrast, under lower overlap ($\sigma{=}20$), methods differ more clearly. Replay-based approaches (e.g., ER) and DMSG achieve stronger performance, while methods relying on structural sparsification (e.g., SSM, SEM) lag behind. Notably, DMSG maintains the most stable forgetting across all $\sigma$, improving from $-14.6$ to $-4.8$, although its advantage in AUC diminishes under heavy overlap.

Overall, these results reinforce the diminishing-leverage pattern observed on CoraFull-CL, while highlighting that the rate of convergence depends on dataset characteristics. In particular, the heterophilous structure of RomanEmpire-CL appears to amplify the effect of overlap, leading to faster saturation and a more pronounced collapse of method differences under strong mixing.

\subsection{Global Mixing}
\label{app:global_mixing}

Tables~\ref{tab:blurry10} and~\ref{tab:blurry30} report the global mixing setting at two mixing proportions $p = 10\%$ and $p = 30\%$. Under this protocol, every task persistently shares a fixed fraction $p$ of its samples with all other tasks throughout the entire stream, simulating a long-tailed background of secondary distributions overlaid on a slowly evolving dominant one.

Raising $p$ benefits every method --- including Bare --- through implicit rehearsal: Bare alone gains $+11.1$, $+7.2$, $+26.3$, and $+13.6$ AA on Arxiv-CL, CoraFull-CL, Reddit-CL, and RomanEmpire-CL respectively when going from $p{=}10\%$ to $p{=}30\%$. Three more nuanced patterns deserve attention.

\paragraph{Saturation dictates how methods cluster.} On Reddit-CL --- by far the easiest dataset for this protocol --- the spread of methods shrinks from $\approx 15$ AA points at $p{=}10\%$ (best SEM $60.2$ vs.~worst MAS$^*$ $45.0$) to $\approx 7$ at $p{=}30\%$ (range $74.0$--$81.4$). Once Bare alone reaches $79.1$, almost no headroom remains for CL machinery to fill, and methods that look strong at $p{=}10\%$ (SEM $+7.4$ over Bare, SSM $+2.6$) collapse to within a few points of the lower bound at $p{=}30\%$.

\paragraph{Detector- and partition-dependent methods stall or regress.} MAS$^*$ fails to meaningfully exceed Bare on any dataset at either $p$ --- most tellingly, on CoraFull-CL at $p{=}30\%$ MAS$^*$ scores $13.9$ versus Bare's $14.4$, and on RomanEmpire-CL the two are tied at $\approx 31.9$. The graph-specific replay methods SSM and SEM exhibit the same pattern on the homophily-dominated, fine-grained datasets (Arxiv-CL, CoraFull-CL): with task identity hidden, their importance/diversity scoring loses its anchor and falls below the simpler ER baseline by margins of $\approx 5$--$30$ points. These observations corroborate the RQ3 analysis in the main text --- methods that internally presume a clean task partition pay the largest price when the partition is dissolved into a persistent background.

\paragraph{Implicit rehearsal does not rescue hard datasets.} On Arxiv-CL (40 classes / 20 latent tasks), the best method at $p{=}30\%$ reaches only $28.1$ --- still $43$ AA points below joint training ($71.6$) --- and beats Bare by a mere $+5.5$, down from $+12.3$ at $p{=}10\%$. The gain delivered by overlapping batches lifts Bare faster than CL machinery can extract additional signal from the buffer, leaving methods regressing toward the lower bound on the very datasets where CL was supposed to help most.

\begin{table}[htbp]
  \centering
  \caption{Performance under global mixing setting (mixing proportion is 10\%). All tasks persistently share 10\% of their data with other tasks. Mean\,$\pm$\,std over 3 runs
    ($\uparrow$ higher is better). Top-2 per column are marked \textbf{1st}~/~\underline{2nd}.}
  \label{tab:blurry10}
  \setlength{\tabcolsep}{1mm}
  \scalebox{0.9}{
  \begin{tabular}{lcccccccc}
    \toprule
    \multirow{2}{*}{Method}
      & \multicolumn{2}{c}{Arxiv-CL}
      & \multicolumn{2}{c}{CoraFull-CL}
      & \multicolumn{2}{c}{Reddit-CL}
      & \multicolumn{2}{c}{RomanEmpire-CL}\\
    \cmidrule(lr){2-3}\cmidrule(lr){4-5}\cmidrule(lr){6-7}\cmidrule(l){8-9}
      & AA\,\%\,$\uparrow$ & AF\,\%\,$\uparrow$
      & AA\,\%\,$\uparrow$ & AF\,\%\,$\uparrow$
      & AA\,\%\,$\uparrow$ & AF\,\%\,$\uparrow$
      & AA\,\%\,$\uparrow$ & AF\,\%\,$\uparrow$ \\
    \midrule
    Bare
      & \vN{11.5}{0.9} & \vN{-80.3}{0.6}
      & \vN{7.2}{1.4}  & \vN{-76.3}{1.0}
      & \vN{52.8}{3.9} & \vN{-42.6}{4.1}
      & \vN{18.2}{2.5} & \vN{-71.4}{2.6} \\
    \midrule
    A-GEM~\cite{chaudhry2018efficient}
      & \vN{15.6}{1.0} & \vN{-76.7}{1.3}
      & \vN{29.2}{3.4} & \vN{-56.4}{3.4}
      & \vN{53.6}{2.1} & \vN{-41.7}{1.9}
      & \vN{23.3}{1.4} & \vN{-66.0}{1.4} \\
    ER~\cite{chaudhry2019rs}
      & \vB{22.0}{1.1} & \vB{-67.0}{2.0}
      & \vB{34.6}{6.3} & \vB{-51.5}{6.3}
      & \vN{55.2}{7.6} & \vN{-40.0}{7.9}
      & \vN{31.5}{1.0} & \vN{-54.5}{1.9} \\
    GSS~\cite{aljundi2019gradient}
      & \vN{19.0}{1.4} & \vN{-72.1}{1.5}
      & \vN{30.6}{8.3} & \vN{-60.5}{7.5}
      & \vN{54.5}{0.8} & \vN{-41.0}{0.8}
      & \vB{34.3}{0.9} & \vB{-52.2}{0.4} \\
    MAS$^*$~\cite{aljundi2019task}
      & \vN{14.1}{0.8} & \vN{-77.7}{0.4}
      & \vN{8.2}{2.9}  & \vN{-76.2}{1.0}
      & \vN{45.0}{5.1} & \vN{-50.4}{6.6}
      & \vN{17.3}{1.8} & \vN{-72.1}{2.0} \\
    SSM~\cite{zhang2022sparsified}
      & \vN{14.6}{1.1} & \vN{-77.6}{1.1}
      & \vN{9.2}{2.4}  & \vN{-78.4}{3.8}
      & \vB{55.4}{4.5} & \vB{-39.3}{4.7}
      & \vN{21.2}{2.1} & \vN{-68.1}{1.7} \\
    SEM~\cite{zhang2023ricci}
      & \vN{14.3}{1.0} & \vN{-78.1}{0.7}
      & \vN{13.0}{5.2} & \vN{-75.4}{7.1}
      & \vA{60.2}{3.1} & \vA{-35.4}{3.1}
      & \vN{22.1}{0.9} & \vN{-67.5}{1.4} \\
    DMSG~\cite{qiao2025towards}
      & \vA{23.8}{0.0} & \vA{-48.5}{1.7}
      & \vA{49.0}{1.1} & \vA{-28.4}{2.7}
      & \vN{53.4}{3.7} & \vN{-39.3}{4.6}
      & \vA{36.6}{1.2} & \vA{-15.4}{3.0} \\
    \bottomrule
  \end{tabular}}
\end{table}

\begin{table}[htbp]
  \centering
  \caption{Performance under global mixing setting (mixing proportion is 30\%). All tasks persistently share 30\% of their data with other tasks. Mean\,$\pm$\,std over 3 runs ($\uparrow$ higher is better). Top-2 per column are marked \textbf{1st}~/~\underline{2nd}.}
  \label{tab:blurry30}
  \setlength{\tabcolsep}{1mm}
  \scalebox{0.9}{
  \begin{tabular}{lcccccccc}
    \toprule
    \multirow{2}{*}{Method}
      & \multicolumn{2}{c}{Arxiv-CL}
      & \multicolumn{2}{c}{CoraFull-CL}
      & \multicolumn{2}{c}{Reddit-CL}
      & \multicolumn{2}{c}{RomanEmpire-CL}\\
    \cmidrule(lr){2-3}\cmidrule(lr){4-5}\cmidrule(lr){6-7}\cmidrule(l){8-9}
      & AA\,\%\,$\uparrow$ & AF\,\%\,$\uparrow$
      & AA\,\%\,$\uparrow$ & AF\,\%\,$\uparrow$
      & AA\,\%\,$\uparrow$ & AF\,\%\,$\uparrow$
      & AA\,\%\,$\uparrow$ & AF\,\%\,$\uparrow$ \\
    \midrule
    Bare
      & \vN{22.6}{1.0} & \vN{-65.0}{2.1}
      & \vN{14.4}{2.8} & \vN{-74.6}{2.3}
      & \vN{79.1}{4.0} & \vN{-14.0}{3.7}
      & \vN{31.8}{3.7} & \vN{-55.4}{4.1} \\
    \midrule
    A-GEM~\cite{chaudhry2018efficient}
      & \vN{25.8}{0.5} & \vN{-61.0}{1.1}
      & \vN{32.6}{6.0} & \vN{-57.8}{5.8}
      & \vN{76.9}{3.6} & \vN{-16.5}{4.0}
      & \vN{34.8}{0.2} & \vN{-51.2}{1.2} \\
    ER~\cite{chaudhry2019rs}
      & \vA{28.1}{0.6} & \vB{-54.9}{2.3}
      & \vB{49.0}{2.2} & \vB{-40.7}{1.1}
      & \vN{74.0}{4.2} & \vN{-19.9}{3.0}
      & \vB{39.0}{1.3} & \vN{-41.9}{1.3} \\
    GSS~\cite{aljundi2019gradient}
      & \vB{27.2}{0.9} & \vN{-58.1}{1.3}
      & \vN{42.4}{0.7} & \vN{-48.1}{1.6}
      & \vB{81.3}{3.8} & \vA{-11.1}{4.8}
      & \vA{43.7}{3.1} & \vB{-36.5}{4.7} \\
    MAS$^*$~\cite{aljundi2019task}
      & \vN{22.5}{1.9} & \vN{-64.7}{2.8}
      & \vN{13.9}{3.1} & \vN{-75.3}{2.0}
      & \vN{78.4}{2.3} & \vN{-14.6}{2.7}
      & \vN{31.9}{3.8} & \vN{-55.1}{4.1} \\
    SSM~\cite{zhang2022sparsified}
      & \vN{24.0}{0.5} & \vN{-63.4}{1.4}
      & \vN{20.2}{4.2} & \vN{-71.0}{4.5}
      & \vN{79.0}{4.4} & \vN{-13.9}{3.6}
      & \vN{31.0}{2.5} & \vN{-55.0}{3.3} \\
    SEM~\cite{zhang2023ricci}
      & \vN{23.6}{0.1} & \vN{-64.0}{0.9}
      & \vN{21.5}{5.4} & \vN{-69.7}{5.1}
      & \vA{81.4}{1.9} & \vB{-12.3}{2.2}
      & \vN{31.9}{3.3} & \vN{-54.4}{2.2} \\
    DMSG~\cite{qiao2025towards}
      & \vN{24.8}{2.0} & \vA{-43.2}{2.3}
      & \vA{52.2}{0.1} & \vA{-23.9}{2.0}
      & \vN{76.6}{6.4} & \vN{-16.2}{6.4}
      & \vN{34.7}{1.6} & \vA{-16.3}{1.7} \\
    \bottomrule
  \end{tabular}}
\end{table}

\subsection{Boundary-Local Mixing}
\label{app:bblurry}

Table~\ref{tab:bblurry} reports the boundary-local mixing setting, where adjacent tasks are mixed only within $K = 5$ batches around each task boundary while the rest of the stream remains task-pure. This regime sits between hard transitions and global mixing: identifiable transition events are preserved, but the boundary itself is softened.

Three observations stand out. First, \textbf{ER is the most consistent top performer}, ranking top-2 on every dataset ($46.5$ on Arxiv-CL, $47.0$ on CoraFull-CL, $53.6$ on Reddit-CL, $55.1$ on RomanEmpire-CL); under this gentle blur, simple uniform reservoir replay matches or beats most of the more elaborate selection schemes. Second, \textbf{DMSG dominates on the fine-grained class-rich datasets} but loses elsewhere --- $63.9$ on CoraFull-CL ($+17$ over ER, the clear best) and $42.6$ on Arxiv-CL (2nd), but only $37.7$ on Reddit-CL and $48.4$ on RomanEmpire-CL --- suggesting that its diversified memory mechanism is most valuable when the underlying class space is large and the dataset is far from saturation. Third, \textbf{MAS$^*$ partially recovers compared to global mixing}: on Arxiv-CL it reaches $41.7$ (vs.~$22.5$ under $p{=}30\%$ global mixing), and on Reddit-CL it actually achieves the best AA in the column ($58.8$). The local mixing window still leaves a sharp enough loss signature for the shift detector to identify, validating the diagnosis in RQ3 that MAS$^*$'s collapse under global mixing stems specifically from the absence of detectable shift events --- not from the presence of overlap per se. Finally, variance is notably high on Reddit-CL for most replay-based methods (std $> 8$ for ER, MAS$^*$, SSM, SEM), reflecting the sensitivity of buffer contents to the specific draw of boundary-window batches when the underlying stream lacks long-term distributional regularity; in contrast, Bare and the gradient-aware GSS remain stable (std $\le 1.5$).

\begin{table}[htbp]
  \centering
  \caption{Performance under the boundary local setting. Adjacent tasks are mixed for $K$ batches at each boundary ($K = 5$ here). Mean\,$\pm$\,std over 3 runs ($\uparrow$ higher is better). Top-2 per column are marked \textbf{1st}~/~\underline{2nd}.}
  \label{tab:bblurry}
  \setlength{\tabcolsep}{1mm}
  \scalebox{0.9}{
  \begin{tabular}{lcccccccc}
    \toprule
    \multirow{2}{*}{Method}
      & \multicolumn{2}{c}{Arxiv-CL}
      & \multicolumn{2}{c}{CoraFull-CL}
      & \multicolumn{2}{c}{Reddit-CL}
      & \multicolumn{2}{c}{RomanEmpire-CL}\\
    \cmidrule(lr){2-3}\cmidrule(lr){4-5}\cmidrule(lr){6-7}\cmidrule(l){8-9}
      & AA\,\%\,$\uparrow$ & AF\,\%\,$\uparrow$
      & AA\,\%\,$\uparrow$ & AF\,\%\,$\uparrow$
      & AA\,\%\,$\uparrow$ & AF\,\%\,$\uparrow$
      & AA\,\%\,$\uparrow$ & AF\,\%\,$\uparrow$ \\
    \midrule
    Bare
      & \vN{24.8}{6.0} & \vN{-66.6}{7.9}
      & \vN{19.9}{2.9} & \vN{-69.1}{3.9}
      & \vN{23.8}{2.7} & \vN{-63.3}{1.5}
      & \vN{31.4}{2.3} & \vN{-58.3}{2.4} \\
    \midrule
    A-GEM~\cite{chaudhry2018efficient}
      & \vN{27.6}{3.1} & \vN{-63.2}{2.4}
      & \vN{37.7}{7.6} & \vN{-51.3}{7.7}
      & \vN{27.8}{7.1} & \vN{-63.8}{6.4}
      & \vN{51.5}{6.3} & \vN{-38.9}{6.7} \\
    ER~\cite{chaudhry2019rs}
      & \vA{46.5}{2.2} & \vB{-39.9}{2.4}
      & \vB{47.0}{1.5} & \vB{-40.1}{2.1}
      & \vB{53.6}{9.2} & \vB{-40.2}{8.5}
      & \vB{55.1}{4.3} & \vN{-31.1}{3.5} \\
    GSS~\cite{aljundi2019gradient}
      & \vN{28.8}{2.1} & \vN{-56.9}{1.9}
      & \vN{42.9}{4.0} & \vN{-48.1}{4.6}
      & \vN{43.5}{1.0} & \vN{-48.2}{2.9}
      & \vA{67.7}{1.6} & \vB{-18.3}{1.6} \\
    MAS$^*$~\cite{aljundi2019task}
      & \vN{41.7}{4.6} & \vN{-46.6}{4.4}
      & \vN{25.9}{3.5} & \vN{-62.4}{4.0}
      & \vA{58.8}{9.9} & \vA{-37.5}{8.9}
      & \vN{45.2}{0.3} & \vN{-42.0}{1.0} \\
    SSM~\cite{zhang2022sparsified}
      & \vN{30.2}{6.1} & \vN{-61.4}{6.0}
      & \vN{21.8}{1.5} & \vN{-67.4}{1.1}
      & \vN{28.6}{8.6} & \vN{-63.4}{9.2}
      & \vN{48.4}{7.1} & \vN{-41.0}{8.0} \\
    SEM~\cite{zhang2023ricci}
      & \vN{29.3}{5.1} & \vN{-63.8}{5.1}
      & \vN{21.8}{2.4} & \vN{-70.7}{2.2}
      & \vN{39.7}{10.7} & \vN{-52.2}{11.8}
      & \vN{49.4}{5.3} & \vN{-39.8}{6.3} \\
    DMSG~\cite{qiao2025towards}
      & \vB{42.6}{0.9} & \vA{-27.0}{1.4}
      & \vA{63.9}{3.5} & \vA{-21.8}{2.9}
      & \vN{37.7}{4.3} & \vN{-53.1}{10.1}
      & \vN{48.4}{2.6} & \vA{-11.3}{2.7} \\
    \bottomrule
  \end{tabular}}
\end{table}

\begin{table}[htbp]
  \centering
  \caption{Effect of doubling the batch size on Arxiv-CL Gaussian mixing ($\sigma{=}60$). Mean$\,\pm\,$std over 3 runs ($\uparrow$ higher is better). Top-2 per column are marked \textbf{1st}~/~\underline{2nd}; $\Delta\text{AUC} = A_{\text{AUC}}(B{=}20) - A_{\text{AUC}}(B{=}10)$.}
  \label{tab:bs_ablation}
  \setlength{\tabcolsep}{1.4mm}
  \scalebox{0.92}{
  \begin{tabular}{lccccc}
    \toprule
    \multirow{2}{*}{Method}
      & \multicolumn{2}{c}{$B{=}10$}
      & \multicolumn{2}{c}{$B{=}20$}
      & \multirow{2}{*}{$\Delta A_{\text{AUC}}$} \\
    \cmidrule(lr){2-3}\cmidrule(lr){4-5}
      & $A_{\text{AUC}}$\,$\uparrow$ & $\text{AF}_{\text{s}}$\,$\uparrow$
      & $A_{\text{AUC}}$\,$\uparrow$ & $\text{AF}_{\text{s}}$\,$\uparrow$
      &  \\
    \midrule
    Bare    & \vN{18.5}{1.5} & \vN{-65.4}{3.1}  & \vN{25.2}{0.9} & \vN{-61.3}{12.3} & $\mathbf{+6.7}$ \\
    A-GEM~\cite{chaudhry2018efficient}
            & \vN{34.1}{1.4} & \vN{-48.6}{4.4}  & \vN{36.9}{1.4} & \vN{-43.5}{0.9}  & $+2.8$ \\
    ER~\cite{chaudhry2019rs}
            & \vB{34.9}{0.8} & \vB{-37.3}{2.8}  & \vB{38.1}{1.3} & \vN{-35.0}{2.3}  & $+3.2$ \\
    GSS~\cite{aljundi2019gradient}
            & \vN{24.2}{3.2} & \vN{-60.4}{5.6}  & \vN{30.9}{2.5} & \vN{-42.9}{3.0}  & $\mathbf{+6.7}$ \\
    MAS$^*$~\cite{aljundi2019task}
            & \vA{38.4}{2.1} & \vA{-22.2}{1.8}  & \vA{43.3}{2.9} & \vA{-12.9}{5.8}  & $+4.9$ \\
    SSM~\cite{zhang2022sparsified}
            & \vN{28.0}{4.3} & \vN{-64.5}{1.9}  & \vN{32.8}{1.4} & \vN{-50.3}{4.8}  & $+4.8$ \\
    SEM~\cite{zhang2023ricci}
            & \vN{26.8}{1.0} & \vN{-59.7}{6.9}  & \vN{32.6}{2.2} & \vN{-46.1}{3.3}  & $+5.8$ \\
    DMSG~\cite{qiao2025towards}
            & \vN{31.6}{0.9} & \vB{-31.9}{2.9}  & \vN{32.8}{0.6} & \vB{-32.5}{3.4}  & $+1.2$ \\
    \bottomrule
  \end{tabular}}
\end{table}

\subsection{Effect of batch size and sampling precision.}
We further examine the impact of batch size on Gaussian mixing, motivated by a practical consideration: when sampling mini-batches according to $\alpha_k(t)$, small batch sizes may introduce discretization effects, where low-probability tasks are under-sampled or entirely omitted due to finite sampling resolution.

Table~\ref{tab:bs_ablation} compares results with batch sizes $B{=}10$ and $B{=}20$ on Arxiv-CL. Increasing the batch size consistently improves $A_{\text{AUC}}$ across all methods, with gains ranging from $+1.2$ to $+6.7$ points. This effect is particularly pronounced for Bare and GSS, suggesting that a larger batch better approximates the intended mixture distribution and increases effective exposure to multiple tasks within each step.

However, the relative ordering of methods remains largely stable. In particular, replay-based methods (e.g., ER) continue to achieve strong performance, while DMSG exhibits the smallest gain ($+1.2$), indicating that its behavior is less sensitive to sampling granularity. Notably, improvements in forgetting are less consistent: while several methods benefit from reduced forgetting, others (e.g., DMSG) show minimal change, suggesting that batch size primarily affects adaptation rather than retention.

Overall, these results indicate that finite-batch sampling introduces a mild approximation error to the target mixture distribution, but does not alter the qualitative conclusions. Larger batches reduce this effect by providing a more faithful realization of $\alpha_k(t)$, while the relative behavior of methods remains consistent.

\begin{figure}[htbp]
    \centering
    \includegraphics[width=0.95\linewidth]{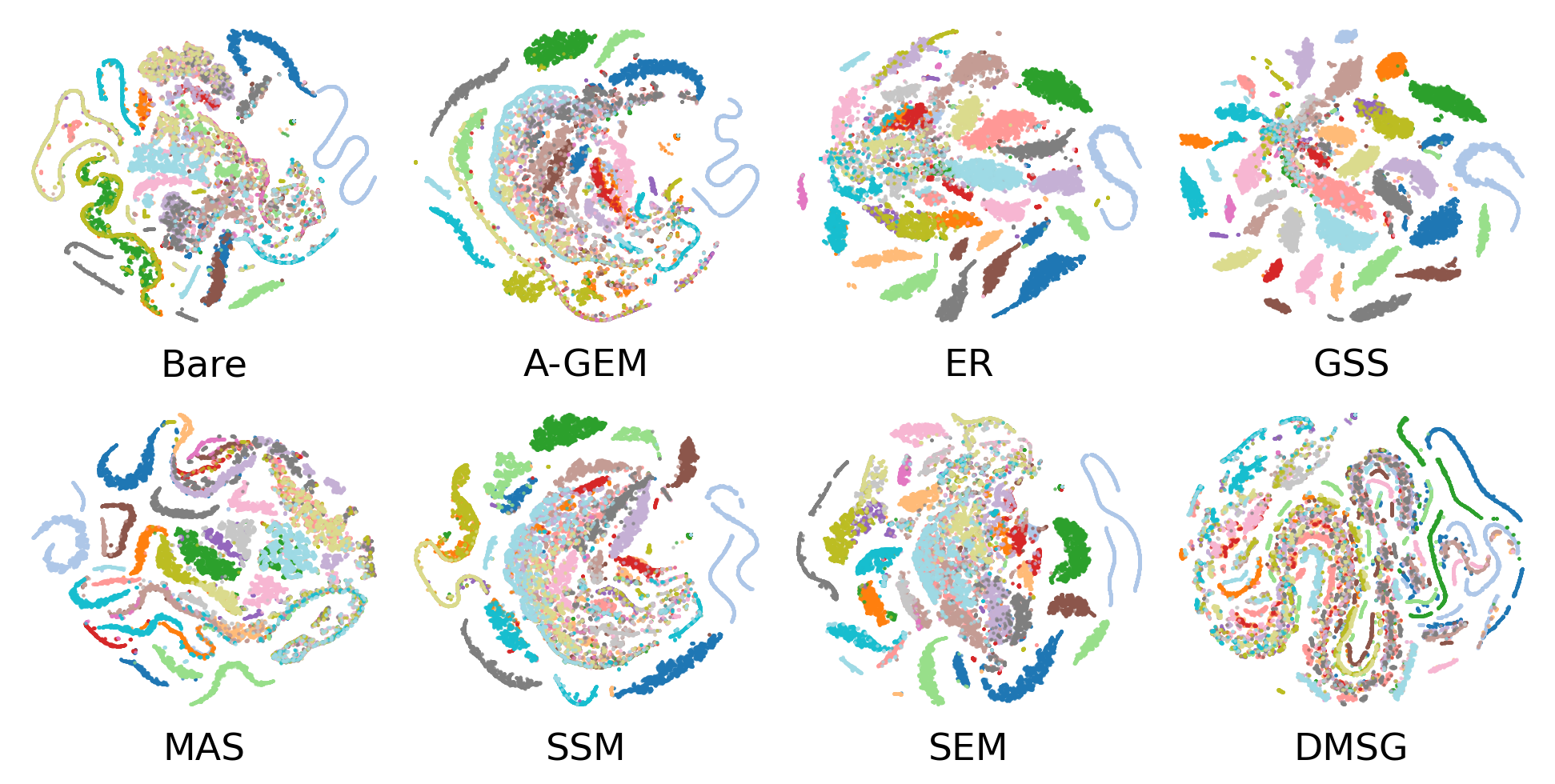}
    \caption{t-SNE visualization of learned node embeddings on Reddit-CL.}
    \label{fig:tsne_reddit}
\end{figure}

\paragraph{Embedding visualization.}
We visualize node representations using t-SNE in Figure~\ref{fig:tsne_reddit}, where colors indicate class labels. DMSG yields more compact and structured clusters, whereas other methods exhibit more dispersed embeddings, reflecting weaker representation quality.

\end{document}